\documentclass[11pt]{article}

\usepackage[final]{acl}

\usepackage{times}
\usepackage{latexsym}

\usepackage[T1]{fontenc}

\usepackage[utf8]{inputenc}

\usepackage{microtype}

\usepackage{inconsolata}

\usepackage{graphicx}
\usepackage{amsmath}
\usepackage{amsfonts} 
\usepackage{booktabs}
\usepackage{multirow}
\usepackage[table]{xcolor}
\usepackage{cleveref}
\usepackage{tcolorbox}
\usepackage{enumitem} 
\newtcolorbox{promptbox}[1]{
  colback=gray!5,
  colframe=gray!50,
  title=#1,
  fonttitle=\bfseries\sffamily,
  fontupper=\small\ttfamily,
  arc=5pt,
  left=2pt,
  right=2pt,
  top=2pt,
  bottom=2pt,
  boxrule=0.5pt
}
\newlist{promptlist}{itemize}{2}
\setlist[promptlist]{label=-, nosep, leftmargin=1.2em, labelsep=0.5em, after=\vspace{0.5em}}
\usepackage[ruled,vlined,linesnumbered]{algorithm2e}
\usepackage{amssymb}
\usepackage{tabularx}
\usepackage{wasysym} 
\usepackage{hyperref}
\usepackage{todonotes} 
\usepackage{lineno}

\title{Beyond Linearization: Attributed Table Graphs for Table Reasoning}

\author{
 {
  \textbf{Yuxiang Wang\textsuperscript{1}}, 
  \textbf{Junhao Gan\textsuperscript{1}}, 
  \textbf{Shengxiang Gao\textsuperscript{1}}, 
  \textbf{Shenghao Ye\textsuperscript{2}}, 
  \textbf{Zhengyi Yang\textsuperscript{3}}, 
  \textbf{Jianzhong Qi\textsuperscript{1,*}} 
  }
  \\
  { \textsuperscript{1}The University of Melbourne \quad \textsuperscript{3}The University of New South Wales} \\
  { \textsuperscript{2}The University of Science and Technology of China} 
  \\
  { \texttt{\{yuxiang.wang8, shengxiang.gao1\}@student.unimelb.edu.au}} \\
  { \texttt{\{junhao.gan, jianzhong.qi\}@unimelb.edu.au}}
}

\begin{document}

\maketitle
\renewcommand{\thefootnote}{*}
\footnotetext{Corresponding author}
\newcommand{\model}{\textsc{TabGR}}
\newcommand{\modelND}{\textsc{TabGR}\textsuperscript{\dag}}
\newcommand{\algo}{\textsc{QG-PPR}}
\newcommand{\wdata}{\texttt{WikiTQ}}
\newcommand{\tdata}{\texttt{TabFact}}
\newcommand{\hdata}{\texttt{HiTab}}
\newcommand{\ldata}{\texttt{SLQA}}
\newcommand{\rdata}{\texttt{RealHiTBench}}

\begin{abstract} 
Table reasoning, a task to answer questions by reasoning over data presented in tables, is an important topic due to the prevalence of knowledge stored in tabular formats. Recent solutions use Large Language Models (LLMs) for their semantic understanding and reasoning capabilities. A common paradigm of such solutions linearizes tables to form plain texts that are served as input to LLMs. 
This paradigm has critical issues. It requires LLMs to infer row-column-cell relations from serialized inputs, makes evidence paths harder to trace, and is subject to the ``lost-in-the-middle'' issue. To address these issues, we propose Table Graph Reasoner (\model), a model that represents tables as an Attributed Table Graph~(ATG) without task-specific training. The ATG explicitly preserves row-column-cell structure while enabling graph-based reasoning over traceable evidence paths. We  further propose a Question-Guided Personalized PageRank (\algo) mechanism to  rerank tabular data and mitigate the lost-in-the-middle issue. Extensive experiments across multiple table reasoning benchmarks show that \model\ consistently outperforms state-of-the-art models by up to $9.7\%$ in accuracy. Our code is available at: \url{https://github.com/yxw-11/TabGR}
\end{abstract}

\section{Introduction}

Table reasoning aims to answer questions by reasoning over data presented in tables. Its importance comes from the prevalence of knowledge stored in tabular formats~\cite{YuCW25}. 

Recent studies leverage the semantic understanding and reasoning capabilities of Large Language Models (LLMs) for table reasoning. They typically follow two  paradigms: (i)~\emph{decomposition-based reasoning}, which generates executable code or conducts iterative symbolic operations to decompose tables, and then reasons over the decomposed sub-tables~\cite{cheng2022binding, ye2023large, wang2024chainoftable, wang2025accurate, YuCW25, ZhangMY25}, and (ii)~\emph{full-table reasoning}, which applies LLMs to reason over full tables directly~\cite{ZhengYJLLSW23,liu2023rethinking,rot}. Both paradigms linearize tables when feeding them into LLMs~\cite{yin2020tabert}, treating tables as flat sequences (e.g., Markdown).

\begin{figure}[t]
    \centering
    \begin{minipage}[c]{1\linewidth}
        \centering
        \includegraphics[width=\linewidth]{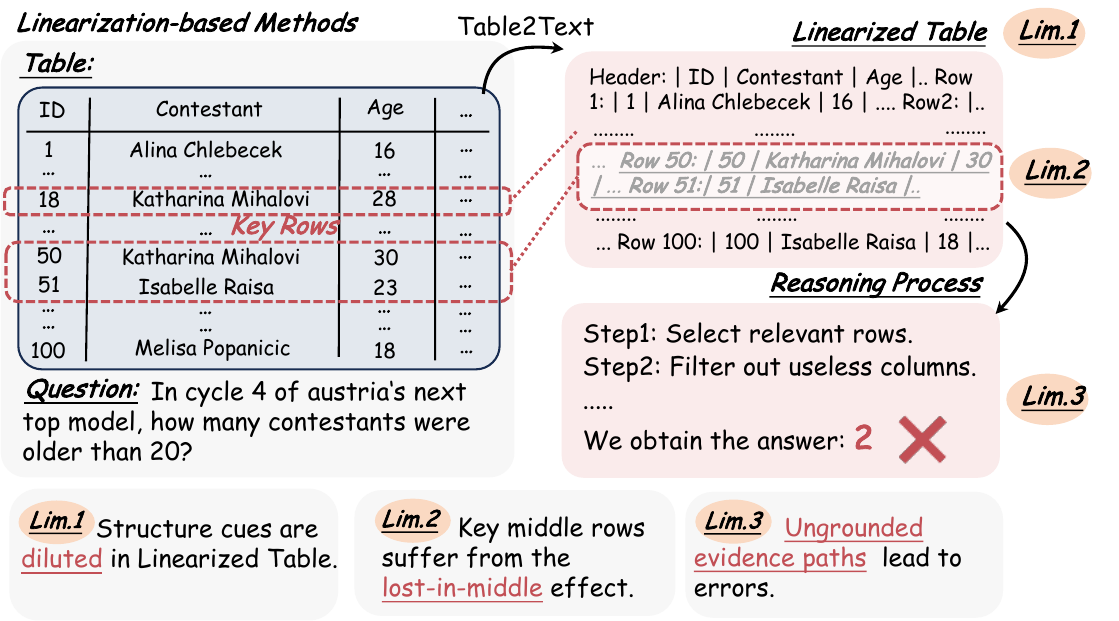}
        \small (a)
    \end{minipage}

        \begin{minipage}[c]{1\linewidth}
        \centering
        \includegraphics[width=\linewidth]{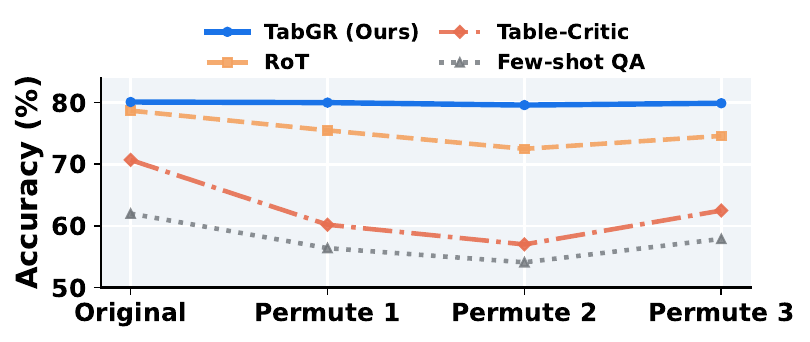}
        \small (b)
    \end{minipage}
    \caption{Motivating example. (a)~Limitations of linearization-based methods. (b)~Performance of different methods on WikiTableQuestions~\cite{pasupat} under random table permutations.}
    \label{fig:limitations}
\end{figure}

While table linearization is simple and has been shown to be effective with existing solutions, we observe a few inherent limitations with such a table representation strategy, as illustrated by Figure~\ref{fig:limitations}. 
 

\textbf{(1) Sequential structure dilution:} Linearized formats encode table structure through textual markup, but structural cues may dilute and are less direct in token sequences than in row-column-cell reasoning units~\cite{tab_llm}. Even with layout cues such as separators and newlines (cf.~Figure~\ref{fig:limitations}a), or richer formats such as HTML or \LaTeX{}, LLMs need to interpret these bindings from sequences, causing grounding errors in challenging tables; examples and comparisons are in Appendix~\ref{app:structural_grounding}. 
Existing structure-aware methods often require architectural adaptation or additional training~\cite{yin2020tabert, abs-2411-02059}, incurring substantial costs.

\textbf{(2) Sensitivity to answer positions:} Linearizing tables into a long sequence is vulnerable to the positions of the answers in table reasoning tasks. LLMs are known to be subject to the ``lost-in-the-middle'' problem~\cite{LiuLHPBPL24}, i.e., they tend to miss information given in the middle of a long input. 
As shown in Figure~\ref{fig:limitations}b, we run two state-of-the-art (SOTA) methods, Table-Critic~\cite{YuCW25} and RoT~\cite{rot}, and a simple few-shot LLM prompting method on three random permutations of the table rows and columns of the WikiTableQuestions dataset~\cite{pasupat}. The table reasoning accuracy fluctuates with each permutation. 

\textbf{(3) Difficult to ground the reasoning process:} Linearizing tables often constrains LLMs to plain-text reasoning, where evidence is serialized text rather than specific rows, columns, or cells.  As existing methods typically perform reasoning (e.g., Chain-of-Thought~\cite{ZhengYJLLSW23}) over serialized tables, their reasoning steps are not explicitly grounded in individual row-column-cell relations, making the supporting evidence harder to trace.

To address these limitations, we propose \underline{Tab}le \underline{G}raph \underline{R}easoner (\model), a novel table reasoning method that introduces an Attributed Table Graph (ATG) for table representation. In the ATG, 
rows and cells of a table are represented as the graph nodes, while the graph edges encode the column relations. This representation allows LLMs to access row-column-cell relations as graph triples. 

Based on the ATG, we propose a Question-Guided Personalized PageRank (\algo)~\cite{JehW03,BrinP12} mechanism to assign question- and structure-aware salience scores to each graph triple by inducing a propagation matrix, and we rerank triples according to their importance. This ensures that crucial information is surfaced to the LLM regardless of its original position in the table. 

Besides, graph-based table reasoning allows LLMs to reason over explicit structural relations by identifying salient subgraphs, tracing reasoning paths, and revealing which row-column-cell connections support the final prediction, making the evidence path easier to inspect.

Our contributions are summarized as follows:

(1)~We propose \model, a table reasoning model that requires no task-specific training and represents tables as an Attributed Table Graph, explicitly preserving the row-column-cell structure for more effective reasoning. 

(2)~We introduce a Question-Guided Personalized PageRank mechanism that assigns salience scores to table (i.e., graph) triples, enabling question- and structure-aware triple reranking and hence improving  robustness to answer position in tables (a.k.a. the ``lost-in-the-middle'' issue).

(3)~We transform table reasoning into graph-based reasoning, enabling LLMs to produce reasoning paths grounded in row-column-cell triples, making the supporting evidence easier to trace.

(4)~Extensive experiments on Table Question Answering (TableQA) and Table Fact Verification (TableFV) benchmarks demonstrate that \model\ not only consistently outperforms SOTA methods in accuracy but also exhibits stronger robustness to permutations of table contents.

\begin{figure*}[t]
\centering
\includegraphics[width = 1\linewidth]{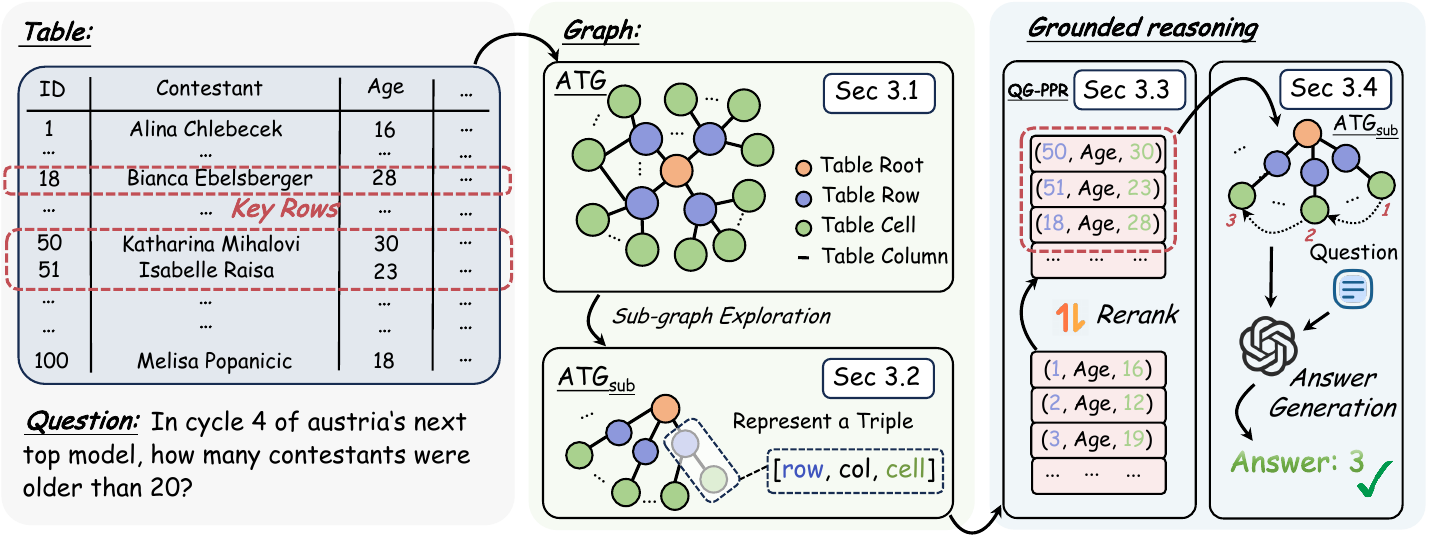}
\caption{Overview of \model, a graph-based framework for table reasoning.
The input table is first transformed into an Attributed Table Graph (ATG),
which supports Question-Guided Personalized PageRank (\algo) for estimating the salience of table triples
and reranking evidence.
The reranked evidence then guides an LLM to perform reasoning over the ATG,
producing a traceable evidence path grounded in table triples.
}
\label{fig:overview}
\end{figure*}

\section{Related Work}
\label{sec:relate-work}
Table reasoning requires models to comprehend both the content and structural information of tables, to support  tasks such as TableQA and TableFV. Existing methods differ primarily in how they represent tables and reason over the  represented format.

\paragraph{Linearization-based Table Reasoning} Most existing methods for TableQA and TableFV follow a ``linearize-then-reason'' paradigm, converting tables into linearized sequences (e.g., in Markdown format) before feeding them into LLMs. The methods can be further categorized  into two groups.

\underline{Decomposition-based reasoning methods}  
generate executable code or symbolic operations to decompose tables and questions. For example, Dater~\citep{ye2023large} decomposes tables and questions via LLM-generated programs. Binder~\cite{cheng2022binding} iteratively refines executable programs with LLM feedback. TableRAG~\cite{ChenME00CFLLP24} generates programs to retrieve relevant table cells and columns before reasoning. Chain-of-Table~\cite{wang2024chainoftable} performs step-wise reasoning by iteratively applying predefined symbolic operations to derive sub-tables. Table-Critic~\cite{YuCW25} improves this pipeline by introducing intermediate critiques to refine each reasoning step. This line of work can reduce long-context exposure through execution, while \model\ instead explicitly keeps table structure during reasoning.

\underline{Full-table reasoning methods} utilize LLMs to reason over full tables. For example, TaCo~\cite{ZhengYJLLSW23} performs Chain-of-Thought (CoT)~\cite{COT} reasoning directly over fully linearized tables. RoT~\cite{rot} improves CoT prompting by performing row-wise traversal reasoning using fully linearized tables. 

Linearization-based table representations are simple and intuitive. However, they encode row-column-cell bindings through serialized markup and create long input contexts, where LLMs may struggle to locate answers hidden in the middle (i.e., the lost-in-the-middle issue~\citep{LiuLHPBPL24}).

\paragraph{Graph Modeling for Tabular Tasks}
Graph structures have been explored to capture dependencies in tabular data. For example, T-RAG~\cite{zou2025gtr} employs hypergraphs for \emph{table retrieval} in multi-table settings, while ODYSSEY~\cite{AgarwalDS25} utilizes hybrid graphs to link tabular cells with external unstructured text for \emph{multi-modal linking}. These methods still rely on \emph{linearized tables} at inference time. GraphOTTER~\cite{LiHLXXL25} targets \emph{complex table layout understanding} by modeling irregular tables as undirected graphs and performs predefined reasoning actions to traverse the graphs. Its graph representation primarily captures local connections through cell values and their positional indices,
lacking explicit global row-column-cell structure, while predefined reasoning actions may limit generalizability. Tunes~\cite{nguyen2025} adopts an entity-centric framework by using LLMs to identify a primary key and infer entity–attribute relations before constructing a graph. Retrieval and reasoning operate at the entity level through hybrid search, where queries are matched to pre-identified entity nodes that are defined by table structure rather than conditioned on the input question, which may limit flexibility across diverse queries.

Different from these methods, \model\ represents tables as an ATG, which maintains row–column–cell connections and enables a \algo\ mechanism. By deriving salience scores from the ATG rather than plain-text, \model\  can prioritize critical triples regardless of their original positions, thereby mitigating the lost-in-the-middle issues. Meanwhile, \model\ creates reasoning paths aligned with table topology, making the supporting evidence easier to trace.

\section{\model}

We consider an input table $\mathcal{T}$ and an input question $\mathcal{Q}$ (in plain-text) that we aim to answer through reasoning over $\mathcal{T}$.   

As Figure~\ref{fig:overview} shows, our Table Graph Reasoner~(\model) first transforms  $\mathcal{T}$ into an Attributed Table Graph (ATG, Section~\ref{subsec:atg}). The ATG explicitly encodes the row-column-cell structure of $\mathcal{T}$ and serves two purposes: 
(1)~It enables LLMs to perform graph-based table decomposition (Section~\ref{subsec:table_decomposition}), allowing \model\ to iteratively zoom in to the question-relevant elements of $\mathcal{T}$. 
(2)~It induces a sparse propagation matrix for a Question-Guided Personalized PageRank~(\algo) mechanism (Section~\ref{subsec:ppr}). By deriving salience scores from the ATG, \model\ reranks graph triples to prioritize the question-critical ones, hence mitigating the ``lost-in-the-middle'' issue. 
The reranked triples are fed into an LLM for final answer generation~(Section~\ref{subsec:answer_generation}). 
This processing pipeline is summarized as Algorithm~\ref{alg:tgr_pipeline} in Appendix~\ref{app:pipeline}.

\subsection{Attributed Table Graph}\label{subsec:atg}

We represent a table $\mathcal{T}$ as an attributed table graph
$\mathcal{G} = (\mathcal{V}, \mathcal{E})$,
where the node set $\mathcal{V}$ contains three types of nodes:
(1)~cell value nodes $c_j^{(k)}$, each corresponding to the $k$-th unique value in column $col_j$;
(2)~row anchor nodes $r_i$, each representing a table row; and
(3)~a root node representing the entire table and connected to all row anchor nodes, mainly used to maintain graph connectivity.
Each cell value node $c_j^{(k)}$ is connected to all row anchor nodes whose rows contain this value in column $col_j$,
and each edge $(r_i, c_j^{(k)}) \in \mathcal{E}$ is annotated with the column header $h_j$. Importantly, identical values appearing in different columns are treated as distinct nodes
and are not merged across columns.

Accordingly, table $\mathcal{T}$ can be viewed as a collection of structured triples defined at the cell level:
\vspace{-1cm}
\begin{center}
\small
\begin{equation}
\langle r_i, h_j, c_{i,j} \rangle, \quad i \in [1, R],~ j \in [1, C],
\end{equation}
\end{center}
assuming $R$ rows and $C$ columns in $\mathcal{T}$,
and $c_{i,j}$ denotes the cell value in row $r_i$ and column $col_j$,
corresponding to the value node $c_j^{(k)}$ in ATG if the value appears
multiple times in the same column.

Constructing $\mathcal{G}$ takes $O(R\cdot C)$ expected time using hashing, where $R\cdot C$ accounts for visiting each cell once in $\mathcal{T}$, and
checking value-node uniqueness takes $O(1)$ time during graph construction. This process and any subsequent graph updates are not question-dependent and can be done offline if $\mathcal{T}$ is given beforehand, where the resulting graph occupies $O(R\cdot C)$ space.

\subsection{ATG-Based Table Decomposition}
\label{subsec:table_decomposition}
\model\ traverses $\mathcal{G}$ to construct a grounded reasoning subgraph that is meant to contain task-relevant structural units (i.e., triples). 

The traversal (i.e., reasoning) process starts by identifying  \emph{anchor triples}. These include (i)~triples whose cell values $c_{i,j}$ \emph{exactly} match a sub-string of the input question $\mathcal{Q}$, and 
(ii)~triples whose edge attribute $h_j$ is considered relevant to $\mathcal{Q}$ by an LLM.



We then iteratively expand the subgraph (i.e. extracted triples), using an LLM \emph{judge} to determine if the current subgraph is sufficient to answer $\mathcal{Q}$. If not, \model\  iteratively selects more columns (multiple columns are allowed for each step), and it adds triples matching the selected columns to the extracted subgraph. 
Otherwise, the traversal is terminated, and the resulting subgraph, denoted by $\mathcal{G}^*$, is passed to the next step for triple reranking. 



\subsection{Question-Guided Personalized PageRank} \label{subsec:ppr}

Next, we present our \algo\ mechanism to rerank triples in $\mathcal{G}^*$, such that the question-critical triples will surface at the top, to catch the attention of LLMs when fed into LLMs for reasoning and answer generation. QG-PPR operates on a triple-level sparse propagation matrix induced from the ATG, where each triple connects only within its row or column, yielding the degree of each triple to be upper bounded by $R + C$. Let $n$ denote the number of triples. Thus, each power iteration of QG-PPR takes $O(n(R + C))$ expected time.

\paragraph{Triple Salience Scores} We compute triple-level salience scores using \algo, which is an adapted Personalized PageRank process, with the following iterative update:
\begin{center}
\small
\begin{equation}
\mathbf{s}^{(t+1)} = \alpha \mathbf{p}_0 + (1-\alpha)\hat{A}^{\top}\mathbf{s}^{(t)},
\end{equation}
\end{center}

where $\mathbf{s}^{(t)} \in \mathbb{R}^{n \times 1}$ denotes the salience scores at iteration $t$,
$\mathbf{p}_0$ is a \emph{question-guided personalization vector} (detailed next),
 $\hat{A}$ is the triple-level sparse propagation matrix (detailed next), and 
$\alpha \in (0,1)$ is the teleport probability~\cite{BrinP12}.

We initialize $\mathbf{s}^{(0)}$ with a uniform distribution and apply power iteration ($K$ times) until convergence. In practice, we run $20$ iterations, which is sufficient to obtain stable salience scores in our experiments.

\paragraph{Personalization Vector} 
We define a personalization vector $\mathbf{p}_0$, which encodes the question-specific importance of each triple and serves as the restart distribution in Personalized PageRank.

Given a triple $\langle r_i, h_j, c_{i,j}\rangle$, we denote by $p_0(i,j)$ the entry in $\mathbf{p}_0$ associated with this triple. 
We initialize $\mathbf{p}_0$ using question-aware signals from both exact string matching and an LLM-based semantic selector.
We construct the key column name set $\mathcal{H}_q$ and key cell value set $\mathcal{C}_q$ by taking the union of
(i)~columns/cell values that exactly match a sub-string of $\mathcal{Q}$ and (ii)~additional columns/cell values selected by the LLM based on semantic relevance. All matched candidates are deduplicated before scoring to avoid repeated counting.

Each triple receives an additive initial score:
\begin{center}
\small
\begin{equation}
p_0(i,j)
= v_{\text{col}} \cdot \mathbb{I}\!\left(h_j \in \mathcal{H}_q\right)
+ v_{\text{val}} \cdot \mathbb{I}\!\left(c_{i,j} \in \mathcal{C}_q\right),
\end{equation}
\end{center}
where $\mathbb{I}(\cdot)$ denotes the indicator function. The entries of the personalization vector $\mathbf{p}_0$ are then normalized
such that $\sum_{i,j} p_0(i,j) = 1$, yielding a valid probability distribution over the triples.
By default, we set $v_{\text{col}}=1.0$ and $v_{\text{val}}=2.0$.

To further down-weight overly frequent cell values and emphasize more informative values,
we incorporate an inverse document frequency (IDF)~\cite{Jones04} term into the value-level contribution $v_{\text{val}}$ by $v_{\text{val}} \cdot \operatorname{IDF}(c_{i,j})$.
Formally, for each cell value $c_{i,j}$, we define:
\vspace{-0.5cm}
\begin{center}
\small
\begin{equation}
\operatorname{IDF}(c_{i,j}) = \log\!\left(1 + \frac{N}{1 + \mathrm{df}(c_{i,j})}\right),
\end{equation}
\end{center}
where $N$ is the total number of rows in $\mathcal{T}$ and $\mathrm{df}(c_{i,j})$
is the number of rows containing $c_{i,j}$.

\paragraph{Sparse Propagation Matrix} 

To incorporate connectivity information between the triples, we induce a weighted triple-level sparse propagation matrix $\hat{A} \in \mathbb{R}^{n \times n}$, projecting row- and column-level structural constraints onto transitions between triples. 
The matrix serves as the transition operator for propagating evidence probability
from question-relevant triples to potential triples.

Two triples $u$ and $v$ are connected if they belong to the same row or share the same column header in  $\mathcal{G}^*$. Their transition weight is defined as:
\vspace{-0.5cm}
\begin{center}
\small
\begin{equation}
w_{u,v} =
\begin{cases}
\dfrac{w_{\text{row}}}{|\mathcal{C}(r_i)|}, & u,v \in \mathcal{C}(r_i), \\[6pt]
\dfrac{w_{\text{col}}}{|\mathcal{C}(col_j)|}, & u,v \in \mathcal{C}(col_j),
\end{cases}
\label{eq:cell_weight}
\end{equation}
\end{center} 
where $\mathcal{C}(r_i)$ and $\mathcal{C}(col_j)$ denote the set of triples from row $r_i$ and column $col_j$, respectively, and $|\cdot|$ represents set cardinality. Parameters $w_{\text{row}}$ and $w_{\text{col}}$ modulate the relative strength of information propagation along the row dimension and column dimension, which satisfies $w_{\text{row}} + w_{\text{col}} = 1$. This constraint guarantees that the sum of transition probabilities from any triple equals 1, preserving numerical stability during the subsequent \algo\ iterations. By tuning these weights, we can flexibly prioritize row content or column content depending on the question type. Further, normalizing by $|\mathcal{C}|$ ensures that the propagation intensity remains invariant to table size and permutation.

\paragraph{Triple Reranking} We use the salience scores $\mathbf{s}$ derived from the \algo\ process to rerank evidence both across rows and within each row. This reranking encourages question-critical triples to be placed at the start of the prompt to be fed into an LLM for reasoning and answer generation, regardless of their positions in the original table $\mathcal{T}$.

For each row $r_i$, we aggregate salience scores of its triples to obtain a row-level importance score:
\vspace{-0.5cm}
\begin{center}
\small
\begin{equation}
\vspace{-0.5em}
S(r_i) = \sum_{j} s_{i,j},
\vspace{-0.3em}
\end{equation}
\end{center}
where $s_{i,j}$ denotes the salience score of the triple $\langle r_i, h_j, c_{i,j} \rangle$.
The rows are then ranked in descending order of $S(r_i)$.

Within each row, the corresponding triples are further ranked
according to their individual salience scores $s_{i,j}$ in descending order,
yielding an intra-row ordering of cell-level evidence.

For \emph{order-sensitive} questions, where the answer depends on the inherent order of the table (e.g., ``who is the first listed player''), the original triple order is preserved. For all other questions, the reranked triples are provided for the downstream reasoning. By prioritizing evidence based on graph-derived salience rather than original position in the table, \model\ effectively mitigates the ``lost-in-the-middle'' effect and achieves robust performance across different table permutations.

\subsection{Answer Generation}
\label{subsec:answer_generation}

The final stage of \model\ utilizes the reranked subgraph to generate both the final answer and explicit reasoning paths. Given question $\mathcal{Q}$ and the reranked subgraph $\mathcal{G}^*_{ranked}$ (in the form of triples), we prompt an LLM to perform Chain-of-Thought (CoT)~\cite{ZhengYJLLSW23} reasoning over $\mathcal{G}^*_{ranked}$.

The LLM is instructed to conduct reasoning by  going through the graph triples. This process can be formulated as follows:
\vspace{-0.5cm}
\begin{center}
\small
\begin{equation}
(P, T, Ans) = \text{LLM}(Q, \mathcal{G}^*_{ranked}),
\vspace{-0.3em}
\end{equation}
\end{center}
where $P = \{t_1, t_2, \dots, t_k\}$ represents the  reasoning path consisting of the sequence of triples $t_i \in \mathcal{G}^*_{ranked}$ that has been used to derive the answer; $T$ denotes the CoT text, which provides the logical derivation grounded by  path $P$, and $Ans$ is the final generated answer.

By running the CoT process over the reranked graph triples rather than a linearized, plain-text table, \model\ effectively decouples reasoning from the table's original sequence. This enables robustness against table permutations and makes  reasoning more traceable, since each step in the path $P$ can be mapped back to triples in $G^{*}_{ranked}$.

\begin{table*}[ht]
\centering
\resizebox{\linewidth}{!}{%
\small
\setlength{\tabcolsep}{4.5pt} 
\renewcommand{\arraystretch}{1.1}

\begin{tabular}{llcccccccc}
\toprule
\multirow{2}{*}{\textbf{Category}} & \multirow{2}{*}{\textbf{Method}} & \multicolumn{2}{c}{\textbf{LLaMA3.3-70B}} & \multicolumn{2}{c}{\textbf{Qwen2.5-72B}} & \multicolumn{2}{c}{\textbf{GPT-4o-mini}} & \multicolumn{2}{c}{\textbf{Average}} \\
\cmidrule(lr){3-4} \cmidrule(lr){5-6} \cmidrule(lr){7-8} \cmidrule(lr){9-10}
& & \texttt{\wdata} & \texttt{\tdata} & \texttt{\wdata} &\texttt{\tdata} & \texttt{\wdata} & \texttt{\tdata} & \texttt{\wdata} & \texttt{\tdata} \\
\midrule
\multirow{7}{*}{Decomp.} 
& Binder~\citep{cheng2022binding}     & 52.2 & 80.5 & 57.0 & 82.2 & 54.8 & 83.3 & 54.7 & 82.0 \\
& GraphOTTER~\citep{LiHLXXL25}     & 55.1 & 71.2 & 55.3 & 76.9 & 50.3 & 67.0 & 53.6 & 71.7 \\
& TableRAG~\citep{ChenME00CFLLP24} & - & - & 59.7 & 72.3 & 55.2 & 69.6 & 57.5 & 71.0 \\
& Dater~\citep{ye2023large}           & 59.5 & 87.6 & 63.0 & 90.0 & 65.8 & 88.3 & 62.8 & 88.6 \\
& Chain-of-Table~\citep{wang2024chainoftable} & 62.1 & 89.9 & 68.3 & 89.7 & 67.5 & 88.9 & 66.0 & 89.5 \\
& Table-Critic~\citep{YuCW25}         & \underline{70.1} & \underline{91.5} & \textbf{77.2} & \underline{92.6} & \underline{73.9} & \underline{91.1} & \underline{73.7} & \underline{91.7} \\
\rowcolor{gray!10}
\multirow{-2}{*}{} & \textbf{\modelND~(ours)}      & \textbf{76.9} & \textbf{93.5} & \underline{76.2} & \textbf{92.8} & \textbf{74.3} & \textbf{91.8} & \textbf{75.8} & \textbf{92.7} \\
\rowcolor{gray!10}
\multicolumn{1}{l}{ }& $\Delta$ (\%) & \multicolumn{1}{c}{\color{red}$\uparrow9.7$}
& \multicolumn{1}{c}{\color{red}$\uparrow2.2$}
& -
& \multicolumn{1}{c}{\color{red}$\uparrow0.2$}
& \multicolumn{1}{c}{\color{red}$\uparrow0.5$}
& \multicolumn{1}{c}{\color{red}$\uparrow0.8$}
& \multicolumn{1}{c}{\color{red}$\uparrow2.8$}
& \multicolumn{1}{c}{\color{red}$\uparrow1.1$} \\

\midrule
\multirow{7}{*}{Full-table} 
& End-to-End QA & 51.1 & 81.0 & 56.6 & 85.0 & 52.6 & 73.5 & 53.4 & 79.8 \\
& Few-Shot QA   & 62.0 & 82.6 & 61.7 & 85.1 & 57.6 & 75.1 & 60.4 & 80.9 \\
& Chain-of-Thought~\citep{COT}        & 72.7 & 91.0 & 73.6 & 92.3 & 72.4 & 90.5 & 72.9 & 91.3 \\
& LLM-Reranker + CoT  & 75.2 & 91.3 & 73.8 & 89.6 & 70.3 & 89.0 & 73.1 & 90.0 \\
& Tunes~\citep{nguyen2025}   & 76.8 & 91.2 & - & - & \underline{75.4} & 90.4 & 76.1 & 90.8 \\
& RoT~\citep{rot}   & \underline{78.7} & \underline{92.6} & \underline{77.3} & \underline{93.6} & 74.2 & \underline{92.0} & \underline{76.7} & \underline{92.7} \\
\rowcolor{gray!10}
\multirow{-2}{*}{} & \textbf{\model~(ours)} & \textbf{80.1} & \textbf{94.4} & \textbf{79.2} & \textbf{94.0} & \textbf{77.1} & \textbf{92.3} & \textbf{78.8} & \textbf{93.6} \\
\rowcolor{gray!10}
\multicolumn{1}{l}{ }& $\Delta$ (\%) & \multicolumn{1}{c}{\color{red}$\uparrow1.8$}
& \multicolumn{1}{c}{\color{red}$\uparrow1.9$}
& \multicolumn{1}{c}{\color{red}$\uparrow2.5$}
& \multicolumn{1}{c}{\color{red}$\uparrow0.4$}
& \multicolumn{1}{c}{\color{red}$\uparrow2.3$}
& \multicolumn{1}{c}{\color{red}$\uparrow0.3$}
& \multicolumn{1}{c}{\color{red}$\uparrow2.7$}
& \multicolumn{1}{c}{\color{red}$\uparrow1.0$} \\

\bottomrule
\end{tabular}
}
\caption{Overall table reasoning performance comparison. \model$^{\dag}$ represents our method  with ATG-based table decomposition and \model\ represents our method with full-graph reasoning. \textbf{Bold} denotes the best performance and \underline{underline} denotes the second-best. $\Delta$ (\%) denotes the relative performance gain of \model\ and \model$^{\dag}$ compared with the best baseline results.}
\label{tab:main_res}
\end{table*}

\begin{table}[ht]
\centering
\small
\setlength{\tabcolsep}{1.6pt}
\renewcommand{\arraystretch}{1.2}
\begin{tabular}{llcccccc} 
\toprule
\multirow{2}{*}{} & \multirow{3}{*}{\textbf{Method}} 
    & \multicolumn{3}{c}{\textbf{\wdata}} 
    & \multicolumn{2}{c}{\textbf{\tdata}} 
    & \multicolumn{1}{c}{\textbf{\ldata}} \\ 
\cmidrule(lr){3-5}\cmidrule(lr){6-7}\cmidrule(lr){8-8}
& &  Small &  Medium  & Large &  Small &  Medium & Large \\
\cmidrule(lr){3-3}\cmidrule(lr){4-4}\cmidrule(lr){5-5}\cmidrule(lr){6-6}\cmidrule(lr){7-7}\cmidrule(lr){8-8}
\multicolumn{2}{l}{\scriptsize \# Questions} & \scriptsize (3,625) & \scriptsize (638) & \scriptsize (81) & \scriptsize (1,942) & \scriptsize (82) & \scriptsize (1,110) \\
\midrule
 & Table-Critic   & 72.1 & 60.8 & \underline{55.6} & 91.7 & 87.8 & 52.3 \\
 & RoT   & \underline{80.6} & \underline{71.4} & 54.3 & \underline{92.7} & \underline{90.2} & \underline{54.1} \\
\midrule
\rowcolor{gray!10}
  & \modelND  & 78.6 & 67.4 & \textbf{70.4} & 93.6 &  92.7 & \textbf{66.7} \\ 
  \rowcolor{gray!10}
   & \textbf{\model}  & \textbf{82.1} & \textbf{72.3} & 65.4 & \textbf{94.5} & \textbf{93.9} & 65.5 \\ 
  \rowcolor{gray!10}
 & $\Delta$ (\%) & \color{red}$\uparrow 1.9$ & \color{red}$\uparrow 1.3$ & \color{red}$\uparrow 26.6$ & \color{red}$\uparrow 1.9$ & \color{red}$\uparrow 4.1$ & \color{red}$\uparrow 23.3$ \\ 
\bottomrule
\end{tabular}
\caption{Table reasoning performance comparison with the best baselines across tables in different size ranges: \textit{small} ($<1$k tokens), \textit{medium} ($1$k--$4$k tokens), and \textit{large} ($>4$k tokens). \wdata\ contains table size from small to large. \tdata\ contains only \textit{small} and \textit{medium} tables, while \ldata\ focuses on real-world \textit{large} tables.}
\label{tab:table_size}
\end{table} 

\section{Experiments}
\label{sec:exp}
\subsection{Experimental Setup}
\paragraph{Datasets} We evaluate our model on five table reasoning benchmarks. We first use two widely adopted datasets, following the recent baseline Table-Critic~\cite{YuCW25}: (i)~\textbf{WikiTableQuestions} (denoted as \wdata)~\cite{pasupat}, a TableQA benchmark with $4,344$ test samples from $421$ tables; and (ii)~\textbf{\tdata}~\cite{ChenWCZWLZW20}, a table fact verification benchmark with $2,024$ test samples from $298$ tables. To further evaluate the generalizability of our method under more complex and large-scale table settings, we additionally conduct experiments on: (iii)~\textbf{\hdata}~\cite{Cheng0WJG0HLZ22}, which features hierarchical tables with multi-level headers and nested structures; (iv)~\textbf{\ldata}~\cite{wang2025large}, a large-table TableQA benchmark; and (v)~\textbf{\rdata}~\cite{realhitbench}, a realistic hierarchical-table benchmark that enables evaluation on complex layouts with table structures recovered from serialized inputs. Dataset details are provided in Appendix~\ref{app:datasets}, and the results on \hdata\ and \rdata\ are reported in Appendix~\ref{app:hitab_res} and ~\ref{app:real_res}, respectively.


\paragraph{Competitors} We compare our method  \model\ (both w/ and w/o table decomposition, denoted as \textbf{\modelND} and \textbf{\model}, respectively) with two categories of baseline methods:  

{(1)~Decomposition-based reasoning:}  This category breaks down tables or questions into sub-components before reasoning, including \textbf{Dater}~\cite{ye2023large}, \textbf{Binder}~\cite{cheng2022binding}, \textbf{TableRAG}~\cite{ChenME00CFLLP24}, \textbf{Chain-of-Table}~\cite{wang2024chainoftable},  \textbf{Table-Critic}~\cite{YuCW25}, and 
\textbf{GraphOTTER}~\cite{LiHLXXL25}, which have been described in Section~\ref{sec:relate-work}.

{(2)~Full-table reasoning:} This category reasons over the full tables. \textbf{End-to-End QA} directly prompts an LLM with linearized full tables and questions. \textbf{Few-shot QA} adds examples (table, question, and the corresponding answer) to the LLM input. \textbf{Chain-of-Thought (CoT)}~\citep{COT} directly prompts an LLM to generate step-by-step reasoning over the linearized table. \textbf{LLM-Reranker + CoT} first prompts the LLM to reorder rows and columns by question relevance, followed by CoT reasoning, providing a strong yet simple reranking baseline. \textbf{Tunes}~\cite{nguyen2025} and \textbf{RoT}~\cite{rot} are also tested, which have been described in Section~\ref{sec:relate-work}. 

\paragraph{Implementation Details} We run experiments with three LLM families of different  sizes, including both open- and closed-source models: 
(i)~GPT-4o-mini~\cite{abs-2410-21276} and GPT-5-mini~\cite{abs-2601-03267}, 
(ii)~LLaMA3.3-70B-Instruct, LLaMA3.1-70B-Instruct, and LLaMA3.1-8B-Instruct~\cite{abs-2407-21783}, and (iii)~Qwen2.5-72B-Instruct~\cite{abs-2412-15115}. LLaMA3.3-70B-Instruct is used as the default backbone model across all methods tested in the same experiment. 

For all baseline methods, we follow their original settings. For our methods, \modelND\ uses ATG-based decomposition, while \model\ reasons over the full ATG. 
We set $\alpha$ to $0.15$ and $0.35$ for the two modes, respectively (detailed hyper-parameter study is included in Appendix~\ref{app:parameter}). For all experiments, we use temperature $0.0$, i.e., greedy decoding, to minimize randomness for result reproducibility. 
The detailed prompts used for our methods are included in Appendix~\ref{app:prompt}.

\paragraph{Evaluation Metrics} For \wdata, \hdata, and \ldata, we use the official \wdata\ evaluator and report accuracy based on exact string matching~\cite{pasupat}. For \tdata, we report binary classification accuracy (true/false).

\subsection{Results}
\paragraph{Accuracy Results} Table~\ref{tab:main_res} reports method performance across different LLM backbones on both datasets. We make the following observations:

(1)~\model\ (with full graph reasoning) consistently outperforms all baseline methods across all three LLMs on both datasets. On average, \model\  achieves $78.8\%$ accuracy on \wdata\ and $93.6\%$ on \tdata, i.e., $2.7\%$ and $1.0\%$ improvements over the strongest baseline, RoT, respectively. McNemar's exact test~\cite{stats_test} confirms significant gains over RoT on both \wdata\ and \tdata\ datasets ($p<10^{-11}$ and $p<0.01$).

(2)~While \modelND\ (with ATG-based table decomposition) is not as accurate as \model\ due to information loss during decomposition, it outperforms all decomposition-based baselines and is on-par with the strongest full-table reasoning baseline, RoT. 
On average, \modelND\ outperforms the best decomposition-based method, Table-Critic, by $2.8\%$ and $1.1\%$ on the two datasets, respectively. 

To further examine the usefulness of \modelND, we break down the tables in the three datasets by size and report method performance over different table size categories in  
Table~\ref{tab:table_size}. We focus on \model, \modelND, and the two best baselines Table-Critic and RoT (same below). 

Overall, larger tables are more challenging, and the different methods achieve lower accuracy over larger tables. \model\ is the most accurate over small or medium tables, while \modelND\ excels over large tables, demonstrating its effectiveness in identifying the most relevant subgraph for answer generation. The performance gap between our methods and the baseline methods is much larger on the large tables, emphasizing the ``lost-in-the-middle'' issue which is addressed by our \algo\ triple reranking. In contrast, small tables yield the best performance for all methods due to low noise and short context. Notably, \modelND\ performs even better on large tables than on medium-sized ones for \wdata, suggesting that ATG-based decomposition helps large-table reasoning by extracting a question-relevant subgraph from many irrelevant triples. Thus, \model\ is recommended for small and medium tables, while \modelND\ is preferable for large tables or tight context budgets.

\begin{table}[t]
\centering 
\small
\setlength{\tabcolsep}{2pt}
\renewcommand{\arraystretch}{1.1}

\begin{tabular}{llcccc}
\toprule
\multirow{2}{*}{} & \multirow{2}{*}{\textbf{Method}} 
    & \multicolumn{2}{c}{\textbf{Shuffled \wdata}} 
    & \multicolumn{2}{c}{\textbf{Shuffled \tdata}} \\
\cmidrule(lr){3-4}\cmidrule(lr){5-6}
&  & Row  & Row \& Col  &  Row  & Row \& Col \\
\midrule

 & Table-Critic   & $\downarrow 14.1$ & $\downarrow 18.7$ & $\downarrow 9.7$ & $\downarrow 14.0$\\

  & RoT  & \underline{$\downarrow 4.1$} & \underline{$\downarrow 6.2$} & \underline{$\downarrow 0.9$} & \underline{$\downarrow 1.1 $}
  \\
\midrule
  \rowcolor{gray!10}
  & \modelND  & \color{blue}$\downarrow 0.9$ & \color{blue}$\downarrow 1.3$ & \color{blue}$\downarrow 0.5$ & \color{blue}$\downarrow 0.7$  \\

  \rowcolor{gray!10}
   & \textbf{\model} & \color{blue}$\downarrow 0.1$ & \color{blue}$\downarrow 0.5$ & \color{blue}$\downarrow 0.6$ & \color{blue}$\downarrow 0.7$
  \\

\bottomrule
\end{tabular}
\caption{Average relative accuracy changes (\%), compared with accuracy on the original datasets.}
\label{tab:table_shuflle}
\end{table}

\paragraph{Robustness Analysis}
To evaluate robustness against answer location in tables, we shuffle table rows and columns with three random seeds and report the average performance drop from the original tables. We identify $1,193$ ($27.46$\%) questions in \wdata\ and $642$ ($31.72$\%) in \tdata\ as order-sensitive and retain their original order.
The remaining order-insensitive questions are shuffled for robustness evaluation.

As shown in Table~\ref{tab:table_shuflle}, the two baseline models report substantial performance drops when the tables are shuffled, possibly because they are tuned for the original datasets. Our \model\ methods exhibit superior robustness, with only marginal accuracy drops, as our \algo\ mechanism prioritizes the question-relevant triples regardless of their initial positions in the tables. The slight drops can be explained by random effects of shuffling. Also, when triples receive identical salience scores, their relative order follows  the (shuffled) table order.

\begin{table}[t]
\centering
\small
\setlength{\tabcolsep}{2.3pt}
\renewcommand{\arraystretch}{1.12}
\begin{tabular}{@{}lllcc@{}}
\toprule
\textbf{Dataset} & \textbf{Category} & \textbf{Method}
& \textbf{Input} & \textbf{Output} \\
\midrule

\multirow{5}{*}{\raisebox{-0.5ex}{\wdata}}
& \multirow{2}{*}{Decomp.}
& Table-Critic
& 31,192
& 875 \\

& & \modelND
& \shortstack{5,063\\{\scriptsize(0.16$\times$)}}
& \shortstack{285\\{\scriptsize(0.33$\times$)}} \\

\cmidrule(lr){2-5}

& \multirow{3}{*}{\raisebox{-0.8ex}{Full-table}}
& RoT
& 2,013
& 460 \\

& & \model\ w/o \algo
& \shortstack{2,500\\{\scriptsize(1.24$\times$)}}
& \shortstack{205\\{\scriptsize(0.45$\times$)}} \\

& & \model
& \shortstack{4,739\\{\scriptsize(2.35$\times$)}}
& \shortstack{223\\{\scriptsize(0.48$\times$)}} \\

\midrule

\multirow{5}{*}{\raisebox{-0.5ex}{\tdata}}
& \multirow{2}{*}{Decomp.}
& Table-Critic
& 30,681
& 1,008 \\

& & \modelND
& \shortstack{4,089\\{\scriptsize(0.13$\times$)}}
& \shortstack{245\\{\scriptsize(0.24$\times$)}} \\

\cmidrule(lr){2-5}

& \multirow{3}{*}{\raisebox{-0.8ex}{Full-table}}
& RoT
& 1,520
& 249 \\

& & \model\ w/o \algo
& \shortstack{1,998\\{\scriptsize(1.31$\times$)}}
& \shortstack{172\\{\scriptsize(0.69$\times$)}} \\

& & \model
& \shortstack{3,665\\{\scriptsize(2.41$\times$)}}
& \shortstack{207\\{\scriptsize(0.83$\times$)}} \\

\bottomrule
\end{tabular}

\caption{Token costs per question. Ratios in parentheses are computed
against Table-Critic for decomposition-based methods and against RoT
for full-table methods.}
\label{tab:cost}
\end{table}

\paragraph{Efficiency Analysis} 
Table~\ref{tab:cost} reports the LLM calling costs in terms of the number of input/output tokens per question. \modelND\ significantly reduces both input/output tokens compared to the best decomposition-based baseline Table-Critic. For example, on \wdata, \modelND\ uses $5,063$ input tokens per question, only $0.16\times$ Table-Critic. This reduction is primarily attributed to ATG-based subgraph extraction, which prunes irrelevant table content before answer generation.

For full-table reasoning, \model\ requires more input tokens than RoT does, mainly because \algo\ constructs question-guided structural evidence through its personalization step. The additional input cost supports stronger structural grounding and robustness, while \model\ still reduces output tokens compared with RoT.  
For efficiency-sensitive settings, \model\ w/o \algo\ provides a \underline{lightweight alternative} that keeps the ATG representation but disables question-guided reranking, using substantially fewer input tokens than full \model\ and fewer output tokens than RoT.
As shown in Table~\ref{tab:ablation}, this variant incurs only $0.8\%$ and $0.6\%$ relative accuracy drops on \wdata\ and \tdata, respectively, and remains competitive with the strongest full-table baseline.

In Appendix~\ref{app:atg_construct}, we further compare the number of LLM calls of different methods, and report \model's ATG construction time and memory consumption across datasets. The results on \ldata\ further demonstrate the scalability of \model.


\begin{table}[t]
\centering
\small
\setlength{\tabcolsep}{4pt}
\begin{tabular}{lcc}
\toprule
\textbf{Method} & \textbf{\wdata} & \textbf{\tdata} \\
\midrule
\multicolumn{3}{l}{\textit{Full-table reasoning}} \\
\model & \textbf{80.1} & \textbf{94.4} \\
\quad w/o \algo & 79.5 & 93.8 \\
\quad w/o Semantic relevance & 77.6 & 92.1 \\
\quad w/o ATG (Linearized CoT) & 72.7 & 91.0 \\
\midrule
\multicolumn{3}{l}{\textit{Decomposition reasoning}} \\
\modelND & \textbf{76.9} & \textbf{93.5} \\
\quad w/o \algo & 76.0 & 93.2 \\
\quad w/o Semantic relevance & 74.6 & 92.4 \\
\quad w/o ATG expansion & 73.7 & 92.0 \\
\quad w/o ATG (Linearized decomp.) & 69.0 & 88.1 \\
\bottomrule
\end{tabular}
\caption{Ablation study results. $\Delta$ (\%) indicates the relative performance drop when a module is removed.}
\label{tab:ablation}
\end{table}


\paragraph{Generalization Results}
We further evaluate \model\ under more challenging table structures and different backbone LLMs. On \hdata, which contains hierarchical tables with multi-level headers and nested structures, \model\ achieves $74.3$\% accuracy, outperforming GraphOTTER at $72.7$\%. On \rdata, we use TreeThinker~\cite{realhitbench} to recover the hierarchical row and column structures from serialized tables. To isolate the effect of table representation and reasoning, both \model\ and RoT use exactly the same recovered structure. Under this controlled setting, \model\ improves the overall accuracy from $53.12$\% to $54.11$\%, and the accuracy on structure comprehension from $64.46$\% to $66.92$\%. These results show that the gains of \model\ extend to more complex table structures and cannot be attributed solely to upstream structure recovery.

We also examine the effect of the backbone LLM in Appendix~\ref{app:across_llms}, where \model\ consistently outperforms strong baselines with additional LLMs of different sizes. Beyond the datasets introduced in Section~\ref{sec:exp}, Appendix~\ref{app:tablebench} evaluates \model\ on TableBench~\cite{tablebench}, providing further evidence on complex real-world table reasoning.

\paragraph{Ablation Study}
We study the contribution of each component through controlled ablations. ``w/o ATG expansion'' uses only anchor triples, ``w/o \algo'' disables question-guided reranking, ``w/o semantic relevance'' disables the LLM-based semantic selector for identifying additional question-relevant columns and cell values beyond exact matches, and ``w/o ATG'' removes ATG representation and graph operations, reducing \model\ to CoT-style reasoning over linearized tables. Table~\ref{tab:ablation} shows that all ablations degrade accuracy, with removing ATG causing the largest performance drop, confirming the importance of explicit row-column-cell grounding. Semantic relevance selection, iterative subgraph expansion, and question-guided reranking provide additional gains. The small drop of \model\ w/o \algo\ further suggests a cost-sensitive variant under limited token budgets, as discussed in Table~\ref{tab:cost}.

\paragraph{Structural Grounding Analysis.}
We investigate whether ATG's gains stem from explicit modeling of table structure or can be achieved with richer serialized textual representations. We compare Markdown, JSON, HTML, and \LaTeX{} with ATG on \wdata, using the same backbone (LLaMA3.3-70B) and prompting setting.

As shown in Table~\ref{tab:structural_grounding}, HTML performs best among the textual formats with $77.3$\% accuracy, while ATG achieves $80.1$\%. We further analyze structural errors, including selecting incorrect rows or columns, associating values with incorrect positions, aggregating the wrong set of cells, and using evidence from irrelevant columns. These errors are judged by GPT-5.3~\cite{abs-2601-03267} using consistent evaluation criteria across all representations. ATG yields a lower error rate of $14.4$\%, compared with $19.8$\% for HTML.

The results indicate that richer serialization alone does not fully preserve table structure for reasoning, since the LLM must still recover the relations among rows, columns, and values from the input sequence. ATG instead encodes each cell as a triple $\langle r_i, h_j, c_{i,j}\rangle$, making these relations explicit and providing more reliable evidence grounding. A qualitative example is provided in Appendix~\ref{app:structural_grounding}.

\begin{table}[t]
\centering
\small
\setlength{\tabcolsep}{5pt}
\begin{tabular}{lcc}
\toprule
\textbf{Representation}
& \textbf{Accuracy}
& \textbf{Struct. Error (\%) $\downarrow$} \\
\midrule
Markdown      & 75.5 & 21.6 \\
JSON          & 74.2 & 23.1 \\
HTML          & 77.3 & 19.8 \\
\LaTeX{}      & 76.3 & 20.4 \\
ATG (\model)  & \textbf{80.1} & \textbf{14.4} \\
\bottomrule
\end{tabular}
\caption{Comparison of table representations on \wdata. Structural errors include selection, positional, aggregation, and cross-column errors.}
\label{tab:structural_grounding}
\end{table}

\paragraph{Other Results} Additional parameter studies, ablations, and case studies are provided in Appendices~\ref{app:parameter}--\ref{app:frontier_model}.

\section{Conclusion}
We proposed \model, a graph-based table reasoning method that represents tables as Attributed Table Graphs, which explicitly preserve row-column-cell structure and enable table-permutation robustness via a Question-Guided Personalized PageRank mechanism. Through reasoning over the attributed table graphs, \model\ produces traceable evidence paths grounded in table triples. Extensive experiments show that our method outperforms state-of-the-art methods in accuracy and robustness to row and column ordering, while offering practical efficiency trade-offs across different variants. 

\section*{Limitations}
Our method assumes that table structure, including row--column relations and cell correspondences, is available or recoverable before ATG construction. For serialized or irregular tables, an upstream structure-recovery module may be required, and its errors may propagate to downstream reasoning.
Improving robustness to imperfect structure recovery remains an important direction for future work.

Our experiments mainly focus on TableQA and TableFV with objective textual outputs, and do not cover open-ended data analysis. Extending \model\ to multimodal and multilingual table reasoning is left for future work.

\section*{Ethics Statement}
All datasets and models used in this paper are publicly available, and their usage strictly adheres to the corresponding licenses and terms of use.

\section*{Acknowledgments}
This work is in part supported by the Australian Research
Council (ARC) via Discovery Project 
DP240101006. Jianzhong Qi is supported by ARC Future Fellowship FT240100170. 
Junhao Gan is in part supported by ARC DP230102908.

\bibliography{custom}

\clearpage
\newpage
\appendix

\section{Prompts}
\label{app:prompt}

\subsection{Prompts for Graph-Based Retrieval}
\label{app:prompt_retrieval}
The prompt used to select the key columns (i.e., edges) for the initial subgraph is as follows.

\begin{promptbox}{Initial column selection}
You are a table analysis assistant.\\[0.5em]
Given:
\begin{promptlist}
    \item A table title
    \item A natural language question about the table
    \item A list of candidate columns from the table
    \item One sample row from the table
\end{promptlist}
Task: Select the column names that contain the information necessary to answer the question. Only return the column names (split by commas), nothing else.\\[0.5em]
Here are some examples:\\
\texttt{[Examples]} \\[0.5em]
Title: \{title\} \\
Question: \{question\} \\
Candidate Columns: \{candidate\_col\} \\
Sample Row: \{sample\_row\} \\
Answer:
\end{promptbox}

The prompt used to check if the subgraph fetched is  sufficient for question answering is as follows.

\begin{promptbox}{Subgraph sufficiency check}
You are given a TableQA question and multiple candidate reasoning paths.\\[0.5em]
Instructions:
\begin{promptlist}
    \item Use ONLY the candidate reasoning paths.
    \item Treat edges as UNDIRECTED: ``(rowK; Col; X)'' and ``(X; Col; rowK)'' both mean rowK have Col = X.
    \item Decide if the provided paths are sufficient to answer the question.
    \begin{itemize}[label=$\ast$, nosep, leftmargin=1em]
        \item Sufficient: the given candidate paths can deterministically yield the answer.
        \item Insufficient: necessary link(s) or value(s) are missing or ambiguous.
    \end{itemize}
    \item If sufficient: Set Finished: True.
    \item If insufficient: Set Finished: False.
    \item Output exactly and only the following one section. Do NOT add any other text.
\end{promptlist}
Here are some examples:\\
\texttt{[Examples]} \\[0.5em]
Title: \{title\} \\
Question: \{question\} \\
Candidate Reasoning Paths: \{reasoning\_paths\} \\
Finished:
\end{promptbox}

The prompt used to select more edges (i.e., column headers) that are used to retrieve corresponding triples to expand the subgraph is as follows.
\begin{promptbox}{Additional edge selection}
You are given a TableQA question, a set of (possibly incomplete) candidate reasoning paths, and a set of available relations from the same table graph.\\[0.5em]
Goal:
\begin{promptlist}
    \item Select the MINIMAL SUFFICIENT subset of relations from Available Relations which, when combined with the candidate reasoning paths, is enough to answer the question.
\end{promptlist}
Rules:
\begin{promptlist}
    \item Use ONLY the relation(s) in provided available relations. Do NOT invent new facts.
    \item Minimality: choose the fewest relations ($\ge 1$) among available relations that make the answer derivable.
    \item Tie-breaking: if multiple equally minimal subsets exist, preserve the original order in Available Relations (i.e., pick the earliest lines that work).
\end{promptlist}
Output format (STRICT):
\begin{promptlist}
    \item Output exactly ONE line that starts with: SELECTED\_RELATIONS: followed by a Python list literal of strings, e.g. ['relation1', 'relation2'].
    \item Do NOT output anything else.
\end{promptlist}

Here are some examples:\\
\texttt{[Examples]}\\[0.5em]
Title: \{title\} \\
Question: \{question\} \\
Candidate Reasoning Paths: \{reasoning\_paths\} \\
Available Relations: \{available\_relations\} \\
Sample Row: \{sample\_row\} \\
SELECTED\_RELATIONS:
\end{promptbox}

\subsection{Prompts for Graph-Based Answer Generation}
\label{app:prompt_ans}
The prompt used to generate the final answers by both \model\ and \modelND is as follows.
\begin{promptbox}{Answer generation}
You are given a TableQA question and a list of candidate reasoning paths. Given a question, you should think it step by step and then answer the question. Also provide the final reasoning path encloses within \texttt{<paths> ... </paths>}.\\[0.5em]
Rules (strict):
\begin{promptlist}
    \item OUTPUT FORMAT: two sections: \\
    \texttt{<think> <paths> ... </paths> ... </think>} \\
    \texttt{<answer> ... </answer>}
\end{promptlist}

Here are some examples:\\
\texttt{[Examples]}\\[0.5em]

YOUR TURN \\
Title: \{title\} \\
Question: \{question\} \\
Header: \{header\} \\
Table Content: \{reasoning\_paths\} \\
\texttt{<think>} \\
\texttt{<paths>}
\end{promptbox}

\begin{algorithm*}[t]
\caption{Overall pipeline of \modelND}
\label{alg:tgr_pipeline}

\SetKwInOut{Input}{Input}
\SetKwInOut{Output}{Output}

\Input{Table $\mathcal{T}$, question $\mathcal{Q}$}
\Output{Final answer $Ans$, Chain-of-Thought text $T$, and reasoning paths $P$.}

 $\mathcal{G} \leftarrow \text{GraphConstruction}(\mathcal{T})$ \hfill $\triangleright$ \text{Represent $\mathcal{T}$ as structured triples} \\

 $\hat{A} \leftarrow \text{InducePropagationMatrix}(\mathcal{G})$ \hfill $\triangleright$ \text{Triple-level transition operator} \\

 $\mathcal{G}^* \leftarrow \text{InitialExtraction}(\mathcal{Q}, \mathcal{G})$ \hfill $\triangleright$ \text{Exact match and relation selection} \\

 $suf \leftarrow \text{False}, count \leftarrow 0$ \\

 \While{$\text{count} < 3$}{
     $suf \leftarrow \text{Judge}(\mathcal{Q}, \mathcal{G}^*)$ \hfill $\triangleright$ \text{LLM-based subgraph sufficiency check} \\
     \If{$suf = \text{True}$}{
         \textbf{break}
    }
     $\mathcal{G}^* \leftarrow \mathcal{G}^* \cup \text{IterativeExpansion}(\mathcal{Q}, \mathcal{G})$ \hfill $\triangleright$ \text{Select additional edges} \\
     $count \leftarrow count + 1$
}

 $\mathbf{s} \leftarrow \text{\algo}(\mathcal{Q}, \mathcal{G}^*, \hat{A})$ \hfill $\triangleright$ \text{Salience estimation via Eq. (5)} \\

$\mathcal{G}^*_{ranked} \leftarrow \text{Rerank}(\mathcal{G}^*, \mathbf{s})$ \hfill $\triangleright$ \text{Inter-row and intra-row reranking} \\

 $(P, T, Ans) \leftarrow \text{LLMReasoning}(\mathcal{Q}, \mathcal{G}^*_{ranked})$ \hfill $\triangleright$ \text{Generate grounded reasoning path and answer} \\

 \Return $P$, $T$ and $Ans$.
\end{algorithm*}

\section{Overall Method Pipeline}
\label{app:pipeline}
Algorithm~\ref{alg:tgr_pipeline} summarizes the processing pipeline of our \modelND\ method. \model\ shares a similar pipeline, but without Lines 3 to 10. Its $\mathcal{G}^*$ is just $\mathcal{G}$.

\section{Dataset Details}
\label{app:datasets}
Table~\ref{tab:dataset} summarizes the five datasets used in the experiments. All experiments are run with three NVIDIA A100 80 GB GPUs on a cloud GPU server.

\begin{table}[h]
\centering
\resizebox{\linewidth}{!}{%
\small 
\setlength{\tabcolsep}{1pt} 
\begin{tabular}{lcccccccc}
\toprule
\multirow{2}{*}{\textbf{Dataset}} &
\textbf{\#Cols.} & \textbf{\#Rows} & \textbf{\#Toks.} &
\textbf{\#Toks.} & \textbf{\#Toks.} &
\multicolumn{3}{c}{\textbf{\#QA Pairs}} \\
\cmidrule(lr){7-9}
 & \textbf{/Tbl.} & \textbf{/Tbl.} & \textbf{/Tbl.} & \textbf{/Cell} & \textbf{/Ans.} & \textbf{Train} & \textbf{Val.} & \textbf{Test} \\
\midrule
\tdata & 6.3 & 13.5 & 317.9 & 3.7 & 1.0 & 92,283 & 12,792 & 2,024 \\
\midrule
\wdata  & 6.4 & 25.4 & 662.6 & 4.1 & 1.7 & 11,321 & 2,831  & 4,344 \\ 
\midrule
\hdata  & - & - & - & - & - & 7,417 & 1,671  & 1,584 \\ 
\midrule
\rdata & 11.9 & 45.6 & - & - & - &
- & - & 3,752 \\
\midrule
\ldata  & 11.0 & 733.8 & \textbf{9,786.2} & 1.2 & \textbf{4.4} & 1,324 & 239  & 1,110 \\ 
\bottomrule
\end{tabular}
}
\caption{Dataset statistics. `Cols.': Columns; `Toks.': Tokens; `/Tbl.': per table.}
\label{tab:dataset}
\end{table}

\begin{table}[t]
\centering
\small
\begin{tabular}{l ccc cc cc}
\toprule
\multirow{2}{*}{\textbf{Method}} & \multirow{2}{*}{\textbf{$w_{\text{row}}$}} & \multirow{2}{*}{\textbf{$w_{\text{col}}$}} & \multicolumn{2}{c}{\textbf{\wdata}} & \multicolumn{2}{c}{\textbf{\tdata}} \\
\cmidrule(lr){4-5} \cmidrule(lr){6-7}
& & & Acc. & $\Delta$ (\%) & Acc. & $\Delta$ (\%) \\
\midrule
\multirow{4}{*}{\modelND} 
& 0.1 & 0.9 & 75.8 & \textcolor{blue}{$\downarrow$ 1.4} & 92.4 & \textcolor{blue}{$\downarrow$ 1.2} \\
& 0.3 & 0.7 & \textbf{76.9} & - & 92.9 & \textcolor{blue}{$\downarrow$ 0.6} \\
& 0.5 & 0.5 & 76.4 & \textcolor{blue}{$\downarrow$ 0.7} & 93.2 & \textcolor{blue}{$\downarrow$ 0.3} \\
& 0.7 & 0.3 & 76.1 & \textcolor{blue}{$\downarrow$ 1.0} & \textbf{93.5} & - \\
& 0.9 & 0.1 & 74.2 & \textcolor{blue}{$\downarrow$ 3.5} & 92.1 & \textcolor{blue}{$\downarrow$ 1.5} \\
\midrule
\multirow{4}{*}{\model} 
& 0.2 & 0.8 & 78.8 & \textcolor{blue}{$\downarrow$ 1.6} & 93.6 & \textcolor{blue}{$\downarrow$ 0.8} \\
& 0.4 & 0.6 & 79.4 & \textcolor{blue}{$\downarrow$ 0.9} & 94.0 & \textcolor{blue}{$\downarrow$ 0.4} \\
& \textbf{0.6} & \textbf{0.4} & \textbf{80.1} & - & \textbf{94.4} & - \\
& 0.8 & 0.2 & 78.5 & \textcolor{blue}{$\downarrow$ 2.0} & 93.3 & \textcolor{blue}{$\downarrow$ 1.2} \\
\bottomrule
\end{tabular}
\caption{Impact of $w_{\text{row}}$ and $w_{\text{col}}$ ($w_{\text{row}} + w_{\text{col}} = 1$). $\Delta$ indicates the relative performance drop from the best result in each group.}
\label{tab:hyper_w}
\end{table}

\begin{table}[t]
\centering
\small
\begin{tabular}{l c cc cc}
\toprule
\multirow{2}{*}{\textbf{Method}} & \multirow{2}{*}{\textbf{$\alpha$}} & \multicolumn{2}{c}{\textbf{WikiTQ}} & \multicolumn{2}{c}{\textbf{TabFact}} \\
\cmidrule(lr){3-4} \cmidrule(lr){5-6}
& & Acc. & $\Delta$ (\%) & Acc. & $\Delta$ (\%) \\
\midrule
\multirow{3}{*}{\modelND} 
& 0.05 & 75.6 & \color{blue}$\downarrow$ 1.7 & 92.8 & \color{blue}$\downarrow$ 0.7 \\
& \textbf{0.15} & \textbf{76.9} & - & \textbf{93.5} & - \\
& 0.25 & 76.1 & \color{blue}$\downarrow$ 1.0 & 93.1 & \color{blue}$\downarrow$ 0.4 \\
\midrule
\multirow{3}{*}{\model} 
& 0.25 & 79.2 & \color{blue}$\downarrow$ 1.1 & 93.9 & \color{blue}$\downarrow$ 0.5 \\
& \textbf{0.35} & \textbf{80.1} & - & \textbf{94.4} & - \\
& 0.45 & 79.5 & \color{blue}$\downarrow$ 0.7 & 94.0 & \color{blue}$\downarrow$ 0.4 \\
\bottomrule
\end{tabular}
\caption{Impact of the teleport probability $\alpha$. $\Delta$ (\%) indicates the relative performance drop from the best result in each group.}
\label{tab:hyper_alpha}
\end{table}

\section{Hyper-parameter Study}
\label{app:parameter}
\subsection{\texorpdfstring{Impact of $w_{\text{row}}$ and $w_{\text{col}}$}{Study on w\_row and w\_col}}
We first tune the propagation weights $w_{\text{row}}$ and $w_{\text{col}}$ under the constraint $w_{\text{row}} + w_{\text{col}} = 1$. We evaluate a set of representative configurations as shown in Table~\ref{tab:hyper_w}, including row-oriented (larger $w_{\text{row}}$), balanced, and column-oriented (larger $w_{\text{col}}$) configurations. 
The best-performing configuration is selected based on validation performance and is used for results reported in the main experiments.

Across both \texttt{\wdata} and \texttt{\tdata}, \model\ consistently achieves its best performance with $w_{\text{row}}=0.6$ and $w_{\text{col}}=0.4$,
indicating a stable preference for moderately stronger row-wise propagation. This suggests that \model\ often benefits from aggregating multiple attributes within the same row, while still preserving sufficient column-level interactions for cross-row comparison. In contrast, \modelND\ exhibits more dataset-dependent sensitivity to the propagation weights.
On \texttt{\wdata}, which typically requires multi-step reasoning involving cross-row filtering and comparison, a  stronger column-wise propagation is preferred.
On \texttt{\tdata}, which focuses on binary verification and often requires checking the consistency of multiple attributes
within the same row, stronger row-wise propagation proves more effective.

From a reasoning perspective, these trends reflect how the two methods process information.
\model\ performs global reasoning over the entire graph structure, allowing salience signals to propagate broadly and resulting in more stable performance across hyper-parameter settings.
\modelND, by contrast, relies more on iterative reasoning over progressively constructed subgraphs, making its performance more sensitive to how relevance is propagated along row or column dimensions.

\subsection{Impact of $\alpha$}
Next, we investigate the impact of the teleport probability $\alpha$, with results summarized in Table~\ref{tab:hyper_alpha}.

The optimal values of $\alpha$ again vary for the two methods. \model$^\dagger$ achieves peak performance at $\alpha=0.15$, whereas \model\ performs the best at $\alpha=0.35$. This suggests that decomposition-based reasoning, which operates on subgraphs, benefits from a lower restart probability to allow for deeper structural propagation over the graphs. Conversely, reasoning over full tables requires a larger $\alpha$ to maintain focus on question-relevant anchors, since it receives larger graphs.


\subsection{Impact of Number of Iterations in \algo}

Table~\ref{tab:ppr_iteration} reports the impact of the power iteration count, $K$, used by \algo. For both \modelND\ and \model, their accuracy improves slightly from $K=10$ to $K=20$ and remains unchanged at $K=30$. Theoretically, the error of the power method for PPR decreases at a geometric rate of $(1-\alpha)^K$. With our default teleport probability $\alpha=0.15$ for \modelND\ and $\alpha=0.35$ for \model, the residual error terms $(0.85)^{20} \approx 0.038$ and $(0.65)^{20} \approx 0.00018$ are sufficiently small to stabilize the relative ranking of the table triples.

\begin{table}[h]
\centering
\small
\begin{tabular}{lccccc}
\toprule
\multirow{2}{*}{\textbf{Method}} & \multirow{2}{*}{\textbf{$K$}} & \multicolumn{2}{c}{\textbf{WikiTQ}} & \multicolumn{2}{c}{\textbf{TabFact}} \\
\cmidrule(lr){3-4} \cmidrule(lr){5-6}
& & Acc. & $\Delta$ (\%) & Acc. & $\Delta$ (\%) \\
\midrule
\multirow{3}{*}{\modelND} 
& 10 & 76.4 & \color{blue}$\downarrow$ 0.7 & 93.1 & \color{blue}$\downarrow$ 0.4 \\
& \textbf{20} & \textbf{76.9} & - & \textbf{93.5} & - \\
& 30 & 76.9 & 0.0 & 93.5 & 0.0 \\
\midrule
\multirow{3}{*}{\model} 
& 10 & 79.8 & \color{blue}$\downarrow$ 0.4 & 94.2 & \color{blue}$\downarrow$ 0.2 \\
& \textbf{20} & \textbf{80.1} & - & \textbf{94.4} & - \\
& 30 & 80.1 & 0.0 & 94.4 & 0.0 \\
\bottomrule
\end{tabular}
\caption{Impact of the number of iterations, $K$.}
\label{tab:ppr_iteration}
\end{table}

\subsection{Impact of the Initial Scoring Weights}
We also evaluate the impact of the initial scoring weights $v_{\text{col}}$ and $v_{\text{cell}}$,
which affect the question-aware relevance distribution in the personalization vector $\mathbf{p}_0$. 

As Table~\ref{tab:score_study} shows, setting $v_{\text{col}}=1.0$ and $v_{\text{cell}}=2.0$
yields the best accuracy.
This observation aligns with our intuition that column headers guide the relevant table fields, while specific cell values play a more direct role in grounding evidence for reasoning.
Incorporating the IDF term further adjusts $v_{\text{cell}}$ by down-weighting frequent values, helping the methods  emphasize more informative evidence. Overall, the relatively small performance variations across different configurations indicate that our graph-based propagation is robust to the choice of initial scoring weights.

\begin{table}[h]
\centering
\resizebox{\linewidth}{!}{
\small
\begin{tabular}{ccccccc}
\toprule
\multirow{2}{*}{\textbf{$v_{\text{col}}$}} & \multirow{2}{*}{\textbf{$v_{\text{cell}}$}} & \multicolumn{2}{c}{\textbf{\wdata\ (Acc.)}} & \multicolumn{2}{c}{\textbf{\tdata\ (Acc.)}} \\
\cmidrule(lr){3-4} \cmidrule(lr){5-6}
& & \modelND & \model & \modelND & \model \\
\midrule
2.0 & 1.0 & 76.6 & 79.9 & 93.4 & 94.2 \\
1.0 & 1.0 & 76.5 & 79.7 & 93.3 & 94.2 \\
\textbf{1.0} & \textbf{2.0} & \textbf{76.9} & \textbf{80.1} & \textbf{93.5} & \textbf{94.4} \\
\bottomrule
\end{tabular}
}
\caption{Impact of the initial scores $v_{\text{col}}$ and $v_{\text{cell}}$.}
\label{tab:score_study}
\end{table}

\section{Additional Ablation Study}
\label{app:ablation}
We conduct additional ablation studies to further evaluate the effectiveness of \algo.

In Table~\ref{app:shuffle_abl}, we remove \algo\ (``w/o \algo'') and rerun our methods on the shuffled datasets as described earlier. 
Both \model\ and \modelND\ become more vulnerable to the shuffles, recording larger drops in accuracy, e.g., $5.9\%$ vs. $0.5\%$ for \model\ w/o \algo\ and \model\ on \wdata\ with randomly shuffled rows and columns. 

\begin{table}[htbp]
\centering 
\resizebox{\linewidth}{!}{%
\small
\setlength{\tabcolsep}{2pt}
\renewcommand{\arraystretch}{1.2}

\begin{tabular}{llcccc}
\toprule
\multirow{2}{*}{} & \multirow{2}{*}{\textbf{Method}} 
    & \multicolumn{2}{c}{\textbf{Shuffled \wdata}} 
    & \multicolumn{2}{c}{\textbf{Shuffled \tdata}} \\
\cmidrule(lr){3-4}\cmidrule(lr){5-6}
&  & Row  & Row \& Col  &  Row  & Row \& Col \\
\midrule

  & \modelND  & \color{blue}$\downarrow 0.9$ & \color{blue}$\downarrow 1.3$ & \color{blue}$\downarrow 0.5$ & \color{blue}$\downarrow 0.7$  \\

 & \modelND w/o \algo  & \color{blue}$\downarrow 6.0$ & \color{blue}$\downarrow 7.1$ & \color{blue}$\downarrow 1.0$ & \color{blue}$\downarrow 1.2$  \\

   & \model & \color{blue}$\downarrow 0.1$ & \color{blue}$\downarrow 0.5$ & \color{blue}$\downarrow 0.6$ & \color{blue}$\downarrow 0.7$
  \\

 & \model\ w/o \algo & \color{blue}$\downarrow 5.8$ & \color{blue}$\downarrow 5.9$ & \color{blue}$\downarrow 1.2$ & \color{blue}$\downarrow 1.5$
  \\

\bottomrule
\end{tabular}
}
\caption{Impact of \algo\ on permutation robustness.}
\label{app:shuffle_abl}
\end{table}

Table~\ref{app:rerank} reports accuracy results when \algo\ is replaced by naive PageRank (``w/ PageRank''), which uses uniform initialization without a personalization vector, i.e., to  propagate importance without explicitly modeling question relevance.

We further compare against a Few-Shot QA baseline with a row-based reranking strategy.
Few-Shot QA relies on linearized table representations, and the row-based reranking ranks table rows according to LLM-predicted importance scores, with higher-ranked rows placed earlier in the LLM input. This setting allows us to examine whether simple LLM-based reranking is sufficient to improve reasoning performance under a table linearization strategy.

The results show that both alternative reranking strategies result in accuracy drops. This indicates that reranking alone is insufficient to address the limitations of linearized table reasoning, and highlights the importance of explicitly modeling table structure, as enabled by \algo's graph-based representation and question-guided propagation.

\begin{table}[t]
\centering
\resizebox{\linewidth}{!}{%
\small
\begin{tabular}{l cc cc}
\toprule
\multirow{2}{*}{\textbf{Method}} & \multicolumn{2}{c}{\textbf{\wdata}} & \multicolumn{2}{c}{\textbf{\tdata}} \\
\cmidrule(lr){2-3} \cmidrule(lr){4-5}
& Acc. & $\Delta$ & Acc. & $\Delta$ \\
\midrule
\textbf{Few-Shot QA } & \textbf{62.0} & - & \textbf{82.6} & - \\
\quad w/ Row-based Reranking  & 58.9 & \color{blue}$\downarrow5.0$ & 74.3 & \color{blue}$\downarrow10.0$ \\
\textbf{\model} & \textbf{80.1} & - & \textbf{94.4} & - \\
\quad w/ PageRank  & 78.9 & \color{blue}$\downarrow1.5$ & 92.5 & \color{blue}$\downarrow2.0$ \\
\bottomrule
\end{tabular}
}
\caption{Impact of \algo\ on improving accuracy.}
\label{app:rerank}
\end{table}

\section{Efficiency and Scalability Analysis}
\label{app:atg_construct}
\paragraph{ATG construction} Table~\ref{tab:runtime_stats} reports the ATG construction time. We see that it takes less than 5 seconds to construct the ATGs required for the $421$ tables of the \wdata\ dataset, i.e., $0.010$ s per table and it is even faster for \tdata, as the average table size is smaller in \tdata. Even for large-table QA benchmark, \ldata\ (with $27$ tables), it takes only $0.019$ s to construct the ATG per table. Also, the maximum time (Max Time) taken for any table in the datasets is below $0.5$ seconds, showing the high practicality of our method. 

In terms of memory usage, constructing an ATG requires approximately $3$ MB per table for \wdata, $1.4$ MB for \tdata, and $45$ MB for \ldata\ (average of $9,786$ tokens per table). This efficiency comes from representing the graph using a CSR sparse matrix, which is well suited for the highly sparse structure of ATGs. Importantly, ATG construction and updates can be performed fully offline as it is query independent, hence minimizing the risk of out-of-memory (OOM) issues during inference.

\paragraph{QG-PPR} We further study the running time of  QG-PPR matrix computation. On average, each question takes $0.17$ s on \wdata\ and $0.08$ s on \tdata\ for this computation. Even on the large-table benchmark \ldata, the running time is only $2.5$ s per question. These results  demonstrate the efficiency of QG-PPR.

To compare the efficiency of QG-PPR with a simpler LLM-Reranker pipeline, we report the average number of input and output tokens per question in Table~\ref{tab:reranker_cost}. On the \wdata\ dataset, although \model\ uses more input tokens, its number of output tokens is much smaller. Since LLM inference latency is largely dominated by the length of generated outputs, the effective inference time of \model\ is only 22\% of that of the simpler reranking method. Similarly, on \tdata, the inference time of \model\ is only 31\% of that of the LLM-Reranker.

These results show that QG-PPR achieves higher accuracy (shown in Table~\ref{tab:main_res}) while substantially reducing output generation time by avoiding prompting the LLM to fully rerank all table cells. As a result, \model\ is not only more accurate but also more efficient in inference than the simpler LLM-Reranker.

\paragraph{Number of LLM Calls} As shown in Table~\ref{tab:llm_calls}, existing decomposition-based methods typically require many iterative LLM calls per question (up to $100$ calls on \wdata). \modelND\ significantly reduces the number of required LLM calls through ATG-based decomposition. The full-table variant \model\ requires only three LLM calls per question on average, achieving highly efficient inference. 
 

\begin{table}[t]
\resizebox{\linewidth}{!}{%
\centering
\small
\setlength{\tabcolsep}{2pt}
\renewcommand{\arraystretch}{1.2}
\begin{tabular}{lcccc}
\toprule
\textbf{Dataset} & \textbf{\# Questions} & \textbf{Total Time (s)} & \textbf{Avg. Time/T (s)} & \textbf{Max Time (s)} \\
\midrule
\wdata  & 4,344 & 4.28 & 0.010 & 0.36 \\
\tdata & 2,024 & 0.73 & 0.002 & 0.15 \\
\ldata & 1,100 & 0.55 & 0.019 & 0.40 \\
\bottomrule
\end{tabular}
}
\caption{ATG construction times.}
\label{tab:runtime_stats}
\end{table}

\begin{table}[t]
\centering
\resizebox{\linewidth}{!}{%
\small
\begin{tabular}{llcc}
\toprule
\textbf{Dataset} & \textbf{Method} & \textbf{Input Tokens} & \textbf{Output Tokens} \\ \midrule
\multirow{2}{*}{\wdata}  & LLM-Reranker + CoT & 2,520 & 1,032 \\
                         & \textbf{\model} & 4,753 & \textbf{223} \\ \midrule
\multirow{2}{*}{\tdata} & LLM-Reranker + CoT & 1,638 & 677 \\
                         & \textbf{\model} & 3,665 & \textbf{207} \\ \bottomrule
\end{tabular}
}
\caption{Token consumption compared with an LLM-based reranker.}
\label{tab:reranker_cost}
\end{table}

\begin{table}[t]
\centering
\small
\begin{tabular}{lc}
\toprule
\textbf{Method} & \textbf{\# LLM Calls} \\
\midrule
Dater~\cite{ye2023large} & 100 \\
Binder~\cite{cheng2022binding}& 50 \\
Table-Critic~\cite{YuCW25} & $\leq 32$ \\
Chain-of-Table~\cite{wang2024chainoftable} & $\leq 25$ \\
\midrule
\modelND & $\leq 9$ \\
\model & \textbf{3} \\
\bottomrule
\end{tabular}
\caption{Comparison of the upper bound of LLM calls per question across methods on the \wdata\ dataset.}
\label{tab:llm_calls}
\end{table}

\section{Experiments on HiTab}
\label{app:hitab_res}
We further evaluate our method on \hdata\ to evaluate its generalizability. \hdata\ is characterized by its intricate hierarchical headers and nested structures, which pose a significant challenge.

To handle the hierarchical structures in HiTab, \model\ introduces two structural adaptations during triple construction to preserve structural and semantic context. 
First, we flatten multi-level column headers by recursively merging parent and child headers into a single semantic attribute path (e.g., Region-Worker Type-Metric), which is embedded into every triple in the column name $h_j$. 
Second, we adapt $c_{i,j}$ by propagating row-level context by forward-filling merged cells in the leftmost columns, ensuring that each triple captures the complete row hierarchy (e.g., Year-Industry). 

We compare \model\ with recent methods, following their original implementations.
In particular, TableParser~\cite{ZhaoJZHWWFZ23}, GraphOTTER~\cite{LiHLXXL25}, and \model\ are all evaluated using Qwen2-72B-Instruct as the backbone LLM, to ensure consistency with their reported settings.

As shown in Table~\ref{tab:hitab_comparison}, \model\ outperforms the strongest baseline GraphOTTER,
which is also a graph-based reasoning method designed for complex table understanding.
This result further demonstrates the flexibility of \model\ in handling more complex table structures.


\section{Experiments on RealHiTBench}
\label{app:real_res}
Following our TableQA/TableFV setting, we evaluate $2,384$ questions from the three objective-answer categories:
\textit{Fact Checking}, \textit{Numerical Reasoning}, and \textit{Structure Comprehending}.
We exclude \textit{Data Analysis} and \textit{Chart Generation} due to their open-ended or executable outputs.

We use TreeThinker~\cite{realhitbench} to recover hierarchical row- and column-header structures from serialized \LaTeX\ tables.
RoT w/ TreeThinker and \model\ w/ TreeThinker use exactly the same recovered hierarchy, with RoT serializing it as textual input and \model\ converting it into complete row/column paths and graph triples. RoT w/ Markdown directly reasons over the flat Markdown representation.
All methods use LLaMA3.3-70B-Instruct as the backbone.

\begin{table}[t]
\centering
\small
\begin{tabular}{lc}
\toprule
\textbf{Method} & \textbf{Accuracy} \\
\midrule
TableParser~\cite{ZhaoJZHWWFZ23}  & 44.6 \\
GPT-3.5~\cite{ZhaoJZHWWFZ23} & 50.0 \\
code-davinci-002~\cite{CaoCLWF23} & 69.3 \\
GraphOTTER~\cite{LiHLXXL25} & 72.7 \\
\midrule
\textbf{\model~(ours)} & \textbf{74.3} \\ 
\bottomrule
\end{tabular}
\caption{Accuracy results on \hdata.}
\label{tab:hitab_comparison}
\end{table}

\begin{table}[t]
\centering
\resizebox{\linewidth}{!}{%
\small
\begin{tabular}{lcccc}
\toprule
\textbf{Method}
& \textbf{Fact}
& \textbf{Numerical}
& \textbf{Structure}
& \textbf{Overall} \\
& \textbf{Checking}
& \textbf{Reasoning}
& \textbf{Compr.}
& \textbf{Accuracy} \\
\midrule
LLaMA3.3-70B
& 53.08 & 36.58 & 55.81 & 48.20 \\
RoT w/ Markdown
& 49.71 & 41.89 & 61.87 & 49.20 \\
RoT w/ TreeThinker
& 53.99 & 45.91 & 64.46 & 53.12 \\
\model\ w/ TreeThinker
& \textbf{54.81} & \textbf{46.43} & \textbf{66.92} & \textbf{54.11} \\
\bottomrule
\end{tabular}
}
\caption{Performance comparison on \rdata\ using LLaMA3.3-70B-Instruct.}
\label{tab:realhitbench_comparison}
\end{table}

As shown in Table~\ref{tab:realhitbench_comparison},
\model\ achieves the best overall Accuracy of 54.11\%. Compared with RoT using the same TreeThinker-recovered structure, \model\ improves overall accuracy from 53.12\% to 54.11\%, and Structure Comprehending from 64.46\% to 66.92\%. This controlled comparison shows that the improvement is not solely due to upstream structure recovery, but also benefits from \model's graph-based structural representation and reasoning.

\begin{table}[t]
\centering
\small
\setlength{\tabcolsep}{8pt}
\renewcommand{\arraystretch}{1.2}

\begin{tabular}{lcc}
\toprule
\textbf{Method} & \textbf{\wdata}
    & \textbf{\tdata} \\
\midrule

 Table-Critic~\cite{YuCW25}   & 45.5 & 68.9\\

   RoT~\cite{rot} & \underline{63.7} & \underline{74.8}  \\
   \midrule
   \rowcolor{gray!10}
    \modelND  & 56.5 & \textbf{75.5} \\
    \rowcolor{gray!10}
    \model  & \textbf{64.2} & 75.2  \\
\bottomrule
\end{tabular}
\caption{Accuracy comparison using LLaMA3.1-8B.}
\label{tab:small_llm}
\end{table}

\begin{table}[t]
\centering
\resizebox{\linewidth}{!}{%
\small 
\begin{tabular}{lcc}
\toprule
\textbf{Method} & \textbf{\wdata} & \textbf{\tdata} \\
\midrule
TabSQLify~\cite{NahidR24} & 57.0 & 70.7 \\
Chain-of-Table~\cite{wang2024chainoftable} & 62.2 & 85.6 \\
PoTable~\cite{potable}    & 65.6 & 87.1 \\
Tunes~\cite{nguyen2025}             & \underline{75.4} & \underline{90.4} \\
\midrule
\rowcolor{gray!10}
\modelND        & 74.2 & 90.9 \\
\rowcolor{gray!10}
\model          & \textbf{76.5} & \textbf{91.6} \\
\bottomrule
\end{tabular}
}
\caption{Accuracy comparison using LLaMA-3.1-70B.}
\label{tab:llama31}
\end{table}

\begin{table}[t]
\centering
\resizebox{\linewidth}{!}{%
\small
\setlength{\tabcolsep}{8pt}
\renewcommand{\arraystretch}{1.2}

\begin{tabular}{lcc}
\toprule
\textbf{Method} & \textbf{\wdata}
    & \textbf{\tdata} \\
\midrule
Chain-of-Thought~\cite{ZhengYJLLSW23} & 76.4 & \underline{94.5}  \\
   RoT~\cite{rot} & \underline{81.7} & 93.6  \\
   \midrule
   \rowcolor{gray!10}
    \modelND  & 83.8 & 95.2 \\
    \rowcolor{gray!10}
    \model  & \textbf{85.4} & \textbf{95.8}  \\
\bottomrule
\end{tabular}
}
\caption{Accuracy comparison using GPT-5-mini.}
\label{tab:gpt5}
\end{table}

\section{Experiments Across Different LLMs}
\label{app:across_llms}
\paragraph{LLaMA3.1-8B} To study the impact of the size of the backbone LLM, we use a smaller LLM, LLaMA3.1-8B, in addition to the LLMs used in Table~\ref{tab:main_res}. As shown in Table~\ref{tab:small_llm}, the accuracy drops for all methods compared with those using LLaMA3.1-70B, which is expected. Our methods still outperform the best baselines in the respective categories, showing their robustness against LLM size. \modelND\ is now slightly more accurate than \model\ on \tdata, because questions in \tdata\ mainly rely on localized evidence, making decomposition-based reasoning particularly beneficial for the smaller LLM with limited information-filtering capability. 

\paragraph{LLaMA3.1-70B} A few baselines, TabSQLify~\cite{NahidR24}, PoTable~\cite{potable}, and Tunes~\cite{nguyen2025}, have used LLaMA3.1-70B in their original papers.
For consistency, here, we also use LLaMA3.1-70B as the backbone LLM for \model. The results in Table~\ref{tab:llama31} show that \model\ outperforms all baselines with this backbone LLM.

\paragraph{GPT-5-mini} We further evaluate with GPT-5-mini, a more recent model in the GPT series. The results in Table~\ref{tab:gpt5} show that \model\ continues to outperform the strongest baselines when all are powered by GPT-5-mini.

\section{Case Study}
\label{app:case_study}
We conduct a case study comparing the outputs of our \model\ with those of two strong baselines: Table-Critic and RoT.

As shown in Figure~\ref{fig:case_study}, \model\ leverages the ATG structure to precisely identify empty cells, whereas RoT and Table-Critic suffer from structural misinterpretation and hallucination, leading to the incorrect inclusion of Row 1.

\begin{figure*}[ht]
\centering
\begin{tcolorbox}[colback=white, colframe=gray!50, arc=5pt, outer arc=5pt, title=Case Study, fonttitle=\bfseries]
\small
\textbf{Question:} During what time period was there no shirt sponsors? \\
\textbf{Gold Answer:} 1982--1985 \\
\textbf{Table Content:} \\
\scriptsize
\begin{tabularx}{\textwidth}{l|c|l|l|l}
\toprule
\textbf{Year} & \textbf{Kit Manufacturer} & \textbf{Shirt Sponsor} & \textbf{Back of Shirt Sponsor} & \textbf{Short Sponsor} \\ 
\midrule
1977--1978 & [Empty] & National Express & [Empty] & [Empty] \\
\rowcolor{yellow!15} 1982--1985 & Umbro & \textbf{[Empty]} & [Empty] & [Empty] \\
1985--1986 & Umbro & Whitbread & [Empty] & [Empty] \\
1986--1988 & Henson & Duraflex & [Empty] & [Empty] \\
1988--1989 & [Empty] & Gulf Oil & [Empty] & [Empty] \\
\dots & \dots & \dots & \dots & \dots \\
2011--2013 & Errea & Mira Showers & Barr Stadia & Gloucestershire Echo \\
\bottomrule
\end{tabularx}
\vspace{0.5em}

\begin{tabularx}{\textwidth}{l|X}
\toprule
\textbf{Model} & \textbf{Reasoning Process \& Prediction} \\
\midrule
\rowcolor{blue!5} \textbf{\model\ (Ours)} & \textbf{Thought:} To find the time period when there was no shirt sponsor, we need to examine the "Shirt Sponsor" column in the table. The table lists various years and their corresponding shirt sponsors. We are looking for rows where the "Shirt Sponsor" is empty, indicating no sponsor. Upon reviewing the table, we find one row where the "Shirt Sponsor" is empty: \textcolor{blue}{(row2; Shirt Sponsor; )} with the year range \textcolor{blue}{(row2; Year; 1982--1985)}. Therefore, the time period when there was no shirt sponsor is 1982--1985. \\
& \textbf{Reasoning Path:} \texttt{(row2; Shirt Sponsor; )} $\rightarrow$ \texttt{(row2; Year; 1982--1985)} \\
& \textbf{Prediction:} \textbf{1982--1985} \hfill \checkmark \textcolor{green}{\textit{Correct}} \\
\midrule
\textbf{Table-Critic} & \textbf{Thought:} To determine the time period with no shirt sponsor, we need to examine the "shirt sponsor" column in the original table. The rows with no shirt sponsor are: Row 1: 1977--1978 (no shirt sponsor); Row 2: 1982--1985 (no shirt sponsor). However, we also notice that there are other rows with no shirt sponsor: Row 6 is not one of them since it has gulf oil as the shirt sponsor, but row 1 and row 2 are. Given the information, the time period with no shirt sponsor is indeed 1977--1978 and 1982--1985. The Prediction Answer only mentions one of these time periods (1982--1985), but it does not account for the other time period (1977--1978). However, since 1982--1985 is indeed one of the correct time periods, the Prediction Answer is partially correct but incomplete. \\
& \textbf{Analysis:} \textcolor{red}{\textbf{Lost Structural Information and Model Confusion.}} The critic misinterprets the table structure (identifying Row 1 as empty) and incorrectly penalizes the correct partial prediction, leading to an over-corrected redundant answer. \\
& \textbf{Prediction:} 1977--1978, 1982--1985 \hfill \scalebox{1.2}{\lightning} \textcolor{red}{\textit{Incorrect}} \\
\midrule
\textbf{RoT} & \textbf{Thought:} To answer the question, I need to figure out during which time period there was no shirt sponsor based on the given table. Let me look at the table again. The table has several columns: Year, Kit Manufacturer, Shirt Sponsor, Back of Shirt Sponsor, and Short Sponsor. My task is to find the time period when the "Shirt Sponsor" column is blank. Let me go through each row one by one. 1. 1977--1978: The "Shirt Sponsor" column is blank. That's one time period. 2. 1982--1985: The "Shirt Sponsor" column is also blank. That's another time period. 3. 1985--1986: The "Shirt Sponsor" column has a value, "Whitbread"... [omitted rows 4-17 for brevity, but similar logic]... So, the time periods when there was no shirt sponsor are 1977--1978 and 1982--1985. \\
& \textbf{Analysis:} \textcolor{red}{\textbf{Structural Grounding Error.}} The model incorrectly interprets Row 1 as blank, failing to ground "National Express" to its correct column due to the limitations of table linearization. \\
& \textbf{Prediction:} 1977--1978, 1982--1985 \hfill \scalebox{1.2}{\lightning} \textcolor{red}{\textit{Incorrect}}\\
\bottomrule
\end{tabularx}
\end{tcolorbox}
\caption{Case study comparing \model\ with strong baselines: Table-Critic and RoT.}
\label{fig:case_study}
\end{figure*}

\subsection{Grounded and Traceable Reasoning}
\label{app:grounded_reasoning}

We provide qualitative examples to illustrate how ATG triples make the reasoning process more traceable. 
The goal is not to claim that textual CoT are not detailed, but to show that \model\ aligns reasoning steps with concrete table evidence.

\paragraph{Case: Assumption drift.}
Table~\ref{tab:case_assumption_drift} shows an excerpt of the original table. 
For the question ``internationally, Marcos Pizzelli has scored multiple goals in Armenia and what other country?'', the gold answer is ``Cyprus''.

\begin{table}[t]
\centering
\small
\setlength{\tabcolsep}{3pt}
\resizebox{\columnwidth}{!}{%
\begin{tabular}{llll}
\toprule
\textbf{Player} & \textbf{Goals} & \textbf{Country/Venue} & \textbf{Notes} \\
\midrule
Marcos Pizzelli & 1 & Armenia & International goal record \\
Marcos Pizzelli & 1 & Cyprus & International goal record \\
Marcos Pizzelli & 1 & Cyprus & International goal record \\
$\cdots$ & $\cdots$ & $\cdots$ & $\cdots$ \\
\bottomrule
\end{tabular}%
}
\caption{Table excerpt for the assumption-drift case study.}
\label{tab:case_assumption_drift}
\end{table}

A CoT baseline predicts ``Andorra'' after shifting the question from player-level scoring records to a team-level proxy. 
This introduces an unsupported assumption that is ungrounded in the target table evidence.

\model\ instead grounds the reasoning path in triples such as 
$\langle r_i, \textit{player}, \textit{Marcos Pizzelli}\rangle$ and 
$\langle r_i, \textit{country/venue}, \textit{Cyprus}\rangle$. 
It counts countries directly from the selected rows and excludes Armenia as specified by the question. 
The selected triples show that ``Cyprus'' appears multiple times for Marcos Pizzelli, making it the remaining country with multiple goals. 
This example shows that a traceable triple-level path can help localize whether an answer is supported by the table or introduced by an unsupported reasoning shortcut.

\section{Analyses on Structural Grounding}
\label{app:structural_grounding}

The results in Table~\ref{tab:structural_grounding} show that ATG achieves higher accuracy and fewer structural errors than serialized textual table representations. Here, we provide a qualitative comparison to illustrate how explicit row-column-cell modeling helps avoid grounding errors that can still occur with richer textual representations.

\subsection{Case Study of Row-Column-Cell Grounding}
\label{app:case_study_structure}

Table~\ref{tab:case_column_grounding} shows an excerpt of a table for the question ``when was the first departure?'', whose gold answer is ``2004-02-27''.

\begin{table}[t]
\centering
\setlength{\tabcolsep}{3pt}
\resizebox{\columnwidth}{!}{%
\begin{tabular}{lllll}
\toprule
\textbf{isn} & \textbf{name} & \textbf{arrival date} &
\textbf{departure date} & \textbf{notes} \\
\midrule
82 & Rasul Kudayev & -- & 2004-02-27 &
Repatriated to Russia on January 3, 2004... \\
$\cdots$ & $\cdots$ & $\cdots$ & $\cdots$ & $\cdots$ \\
\bottomrule
\end{tabular}%
}
\caption{Table excerpt for the column-grounding case study.}
\label{tab:case_column_grounding}
\end{table}

The \LaTeX{}-based method incorrectly predicts ``January 3, 2004'', using a date from the \textit{notes} column instead of ``2004-02-27'' from the \textit{departure date} column. Although \LaTeX{} provides richer layout information than simpler textual formats, the model must still infer the association between each value and its corresponding column from the serialized input. In contrast, \model\ explicitly represents the relevant evidence as
$\langle r_1, \textit{departure date}, \textit{2004-02-27}\rangle$.
The date ``January 3, 2004'' is separately associated with the
\textit{notes} column, preventing it from being confused with the target departure date. \model\ therefore returns the correct answer, ``2004-02-27''.

This example complements the quantitative results in Table~\ref{tab:structural_grounding}. Richer textual representations can preserve more layout information, but they still require the LLM to recover row-column-cell relations from the input sequence. ATG encodes these relations explicitly, providing more reliable grounding and making the supporting evidence directly traceable.

\section{Comparison with Program-based Methods}
\label{app:program_methods}

Program-based methods can mitigate long-context issues by avoiding full-table processing through Python or SQL execution. 
However, this paradigm may introduce program-generation and execution errors, including wrong column/filter selection and information loss from early filtering or aggregation.

We compare \model\ with representative program-based methods on table understanding tasks. 
Python QA directly generates Python code to answer questions. 
Binder iteratively refines and executes generated programs. 
DATER~\cite{ye2023large} decomposes tables and questions into sub-tables and uses SQL to answer sub-questions. 
TableRAG~\cite{ChenME00CFLLP24} retrieves relevant cells through generated Python programs. 
TabSQLify~\cite{NahidR24} generates SQL queries based on table schemas. 
TALON~\cite{talon} uses a multi-agent framework with code execution.

\begin{table}[t]
\centering
\small
\setlength{\tabcolsep}{5pt}
\begin{tabular}{lcc}
\toprule
\textbf{Method} & \textbf{GPT-4o-mini} & \textbf{Qwen2.5-72B} \\
\midrule
Python QA & 51.2 & 41.9 \\
Binder & 61.4 & 58.5 \\
DATER & 59.8 & 60.2 \\
TableRAG & 55.2 & 59.6 \\
TabSQLify & 56.5 & -- \\
TALON & 71.0 & 69.6 \\
\modelND & 74.3 & 76.2 \\
\model & \textbf{77.1} & \textbf{79.2} \\
\bottomrule
\end{tabular}
\caption{Comparison with representative program-based methods.}
\label{tab:program_methods}
\end{table}

\begin{table}[t]
\centering
\small
\setlength{\tabcolsep}{6pt}
\begin{tabular}{lc}
\toprule
\textbf{Method} & \textbf{ROUGE-L} \\
\midrule
Chain-of-Thought & 38.2 \\
RoT & 44.8 \\
\model & \textbf{45.5} \\
\bottomrule
\end{tabular}
\caption{Additional evaluation on TableBench.}
\label{tab:tablebench}
\end{table}

As shown in Table~\ref{tab:program_methods}, \model\ outperforms the best baseline by $6.1$ points on GPT-4o-mini and $9.6$ points on Qwen2.5-72B. 
Although program-based methods can avoid processing full table sequences, they may break the original table into partial views or introduce program-generation and execution errors. 
In contrast, \model\ and \modelND\ preserve table structure and sufficient context through ATG, leading to stronger performance. 
Program execution is therefore complementary to our work and could be combined with \model\ in future systems.

\section{Additional Evaluation on TableBench}
\label{app:tablebench}

To further evaluate \model\ in more challenging table reasoning settings, we conduct additional experiments on TableBench~\cite{tablebench}. 
TableBench contains complex real-world table reasoning examples and complements the main benchmarks used in Section~\ref{sec:exp}.

As shown in Table~\ref{tab:tablebench}, \model\ achieves the best ROUGE-L among the compared methods, with a modest gain over RoT. 
This suggests that ATG-based reasoning can generalize to more complex table reasoning settings. The smaller margin suggests that TableBench remains challenging and may benefit from stronger hybrid reasoning strategies.

\section{Frontier Model Evaluation and Failure Cases}
\label{app:frontier_model}

To examine whether standard table reasoning benchmarks remain challenging for frontier LLMs, we evaluate Claude Opus 4.7 (via Anthropic API) on the full test sets of \wdata\ and \tdata.
As shown in Table~\ref{tab:frontier_model}, \model\ further improves Claude Opus 4.7 on both datasets, demonstrating that our structured reasoning framework remains beneficial even with a strong frontier LLM. This suggests that these benchmarks remain challenging, particularly for precise filtering, comparison and structural reasoning.

\begin{table}[t]
\centering
\small
\setlength{\tabcolsep}{6pt}
\begin{tabular}{lcc}
\toprule
\textbf{Method} & \textbf{\wdata} & \textbf{\tdata} \\
\midrule
Claude Opus 4.7 & 78.6 & 91.2 \\
\model\ w/ Claude Opus 4.7 & \textbf{80.9} & \textbf{93.7} \\
\bottomrule
\end{tabular}
\caption{Evaluation with Claude Opus 4.7 on \wdata\ and \tdata.}
\label{tab:frontier_model}
\end{table}

\begin{table}[t]
\centering
\small
\setlength{\tabcolsep}{5pt}
\begin{tabular}{lc}
\toprule
\textbf{Model} & \textbf{mpg} \\
\midrule
Fiat 500 1.2 POP & 46 \\
Mini Cooper COUPE 6M 3DR 1.6L Diesel & 52 \\
\bottomrule
\end{tabular}
\caption{Excerpt for the entity-comparison failure case.}
\label{tab:frontier_case_compare}
\end{table}

\begin{table}[t]
\centering
\small
\setlength{\tabcolsep}{5pt}
\begin{tabular}{ll}
\toprule
\textbf{Single} & \textbf{Album} \\
\midrule
$\cdots$ & A World Called You \\
$\cdots$ & A World Called You \\
$\cdots$ & single only \\
$\cdots$ & A World Called You \\
\bottomrule
\end{tabular}
\caption{Excerpt for the counting/filtering failure case.}
\label{tab:frontier_case_count}
\end{table}

\begin{table}[t]
\centering
\small
\setlength{\tabcolsep}{5pt}
\begin{tabular}{llll}
\toprule
\textbf{Year} & \textbf{Name} & \textbf{Year} & \textbf{Name} \\
\midrule
$\cdots$ & $\cdots$ & 1962--63 & Mrs S. Von Wielligh \\
$\cdots$ & $\cdots$ & 1963--64 & Mr F.J. Van Heerden \\
\bottomrule
\end{tabular}
\caption{Excerpt for the structural-lookup failure case.}
\label{tab:frontier_case_lookup}
\end{table}

\paragraph{Failure Case 1: Entity comparison error.}
Table~\ref{tab:frontier_case_compare} shows a WikiTQ example that requires comparing numerical values across entities. 
The question asks which vehicle gets better mpg. 
The gold answer is ``Mini Cooper'', since $52 > 46$, but the frontier model predicts ``Fiat 500''. 
This indicates that even frontier models can make incorrect entity comparisons when reasoning over tabular values.

\paragraph{Failure Case 2: Counting and filtering error.}
Table~\ref{tab:frontier_case_count} shows an example where the question asks how many singles are only a single. 
The gold answer is $18$, while the frontier model predicts $22$. 
The error comes from incorrect filtering on the \textit{Album} column, where entries associated with an album are over-counted as ``single only'' cases.

\paragraph{Failure Case 3: Structural lookup error.}
Table~\ref{tab:frontier_case_lookup} shows an example with repeated column groups. 
The question asks who is listed next after ``Mrs S. Von Wielligh''. 
The gold answer is ``Mr F.J. Van Heerden'', but the frontier model fails to follow the local table order under the repeated column structure. 
This reflects a structural lookup error, where the model does not correctly track adjacency within the intended column group.

\paragraph{Analysis.}
These failure cases show that frontier LLMs can still struggle with table reasoning operations that require faithful grounding to table structure, including numerical comparison, column-based filtering, and local structural lookup. 
These errors are consistent with the motivation of \model, which represents table evidence as structured triples and ranks question-relevant evidence before answer generation.

\end{document}